\documentclass[10pt,twocolumn,letterpaper]{article}

\usepackage[final,applications]{wacv}

\usepackage{amsmath}
\usepackage{amssymb}
\usepackage{graphicx}
\usepackage{booktabs,colortbl}
\usepackage{multirow}
\usepackage{pifont}
\newcommand{\cmark}{\ding{51}}%
\newcommand{\xmark}{\ding{55}}%

\usepackage[pagebackref,breaklinks,colorlinks]{hyperref}

\usepackage[capitalize]{cleveref}
\crefname{section}{Sec.}{Secs.}
\Crefname{section}{Section}{Sections}
\Crefname{table}{Table}{Tables}
\crefname{table}{Tab.}{Tabs.}

\def\wacvPaperID{413} 
\def\confName{WACV}
\def\confYear{2025}

\begin{document}

\title{VG-SSL: Benchmarking Self-supervised Representation Learning Approaches for Visual Geo-localization}


\author{%
  Jiuhong Xiao \\
    New York University\\
  \texttt{jx1190@nyu.edu}
  \and
  Gao Zhu \\
    New York University\\
  \texttt{gz706@nyu.com}
  \and
  Giuseppe Loianno 
  \\
    New York University\\
  \texttt{loianno@nyu.edu}
}

\maketitle

\begin{abstract}

Visual Geo-localization (VG) is a critical research area for identifying geo-locations from visual inputs, particularly in autonomous navigation for robotics and vehicles. Current VG methods often learn feature extractors from geo-labeled images to create dense, geographically relevant representations. Recent advances in Self-Supervised Learning (SSL) have demonstrated its capability to achieve performance on par with supervised techniques with unlabeled images. This study presents a novel \textbf{VG-SSL} framework, designed for versatile integration and benchmarking of diverse SSL methods for representation learning in VG, featuring a unique geo-related pair strategy, \textbf{GeoPair}. Through extensive performance analysis, we adapt SSL techniques to improve VG on datasets from hand-held and car-mounted cameras used in robotics and autonomous vehicles. Our results show that contrastive learning and information maximization methods yield superior geo-specific representation quality, matching or surpassing the performance of state-of-the-art VG techniques. To our knowledge, This is the first benchmarking study of SSL in VG, highlighting its potential in enhancing geo-specific visual representations for robotics and autonomous vehicles. The code is publicly available at \url{https://github.com/arplaboratory/VG-SSL}.

\end{abstract}

\section{Introduction}\label{sec:introduction}
\begin{figure}
    \centering
    \includegraphics[width=\linewidth]{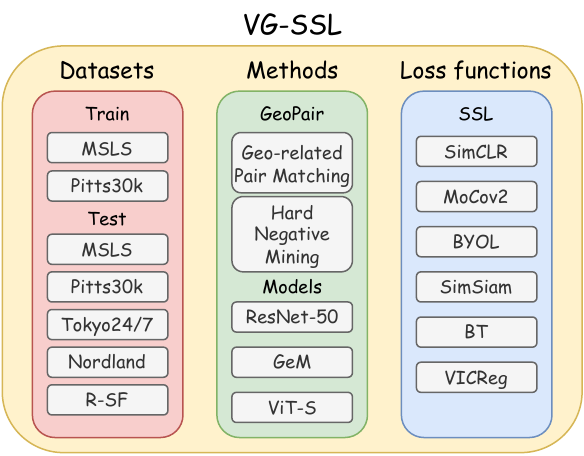}
    \caption{\textbf{VG-SSL} Framework Overview: This framework integrates various Visual Geo-localization (VG) datasets, models, and Self-Supervised Learning (SSL) loss functions for representation learning. It benchmarks VG performance across different SSL strategies trained with the geo-related pair strategy, \textbf{GeoPair}, and offers an in-depth analysis of SSL method settings tailored for geo-specific representation learning.}
    \label{teaser}
\end{figure}

Visual Geo-localization (VG), also known as Visual Place Recognition, is a rapidly growing research area in computer vision~\cite{Berton_CVPR_2022_benchmark, Berton_CVPR_2022_CosPlace, ali2023mixvpr, netvlad} and robotics~\cite{tfvpr, doi:10.1177/0278364919839761, 7989366, xiao2023long}. Its primary goal is to identify geographical locations based on visual features. Most VG methods follow the image retrieval paradigm, building a database of robust, compact representations (e.g., embeddings or histograms of orientation~\cite{SIFT}) from geo-referenced images containing longitude and latitude data. Recent approaches heavily rely on deep neural networks~\cite{lecun2015deep} to create these representations, projecting images into an embedding space. The representation of a query image is then matched to the closest embedding in the database to retrieve geo-referenced information. For accurate matching, the feature extractor must produce representations that are resilient to noise, viewpoint changes, and dynamic objects. This aligns with the principles of representation learning, which focuses on generating robust and informative data representations, with or without specific downstream tasks. This concept has become increasingly important in computer vision, natural language processing~\cite{representation}, and robotics~\cite{SAVIOLO2023}.

Self-Supervised Learning (SSL) techniques recently emerged as popular solutions for representation learning in computer vision. These methods rely on pre-training feature extractors on pretext tasks \cite{pretext} with vast amounts of unlabeled data to learn visual representation. Once pre-training, these models can be fine-tuned by freezing the pre-trained feature extractor and training additional downstream heads for specific tasks. With a pre-trained feature extractor, they can achieve on-par performance with much less labeled data. For example, a common SSL training strategy is contrastive learning \cite{simclr, moco, mocov2}. This approach aims to minimize the embedding distance between augmented image views of the same instance and maximize the distance between views of different instances. 

In pursuit of the shared objective of VG and SSL methods to learn robust and recognizable representations, this study aims to leverage SSL methodologies for VG tasks to assess their capability in learning geo-specific representations. However, relying solely on SSL's data augmentation is insufficient for a comprehensive understanding of real-world geographic relationships, which can be enhanced by geo-labels. To bridge this gap, we introduce a novel pairing strategy that incorporates both data augmentation and geo-labels within SSL methods, aiming to improve the learning of geo-specific representations significantly.
 
Our main contributions can be summarized as follows:
\begin{itemize}
    \item We introduce the novel \textbf{VG-SSL} framework (Fig.~\ref{teaser}) to benchmark self-supervised representation learning methods such as SimCLR, MoCov2, BYOL, SimSiam, Barlow Twins, and VICReg. Validated across multiple large-scale VG datasets, this framework offers a comprehensive view of SSL methods in VG contexts and also includes a unified interface for extending to other SSL approaches.

    \item We propose the \textbf{GeoPair} strategy as a new data pairing strategy for VG tasks. It not only leverages geo-labels to learn and capture geographical relationships as well as employs data augmentation and advanced SSL loss functions to enhance sample efficiency.
    
    \item Extensive evaluations of the proposed framework on robotics and autonomous vehicle VG datasets demonstrate that contrastive learning and information maximization methods effectively learn geo-specific representations with complex spatial relationships of data, achieving performance on par with or superior to current leading VG methods. Further analysis identifies key architectural settings that significantly impact VG performance.
\end{itemize}

To our knowledge, this is the first unified framework for benchmarking SSL methods in VG tasks. It not only lays the foundation for enhancing geo-localization with SSL techniques but also provides an effective training approach for SSL visual foundational models to learn geo-relationships in data.

\section{Related Works}\label{sec:relatedworks}
{\bf Representation Learning for VG.} VG methods aim to extract robust geo-related representations from images to differentiate distinct locations. Traditional VG approaches~\cite{fab-map, bobw, seq-slam, semantic_vpr} rely on handcrafted templates, which are generally less resilient to noise and appearance changes (e.g., illumination). In contrast, recent deep learning-based VG methods~\cite{netvlad, ali2023mixvpr, Berton_CVPR_2022_CosPlace, ZHANG2021107760, ALIBEY2022194, 7410507} leverage deep neural networks~\cite{lecun2015deep} for geo-specific representation learning, incorporating local feature extraction and trainable feature aggregation modules. Popular feature aggregation techniques include MAC~\cite{razavian2016visual}, GeM pooling~\cite{8382272}, and NetVLAD~\cite{netvlad}, with various extensions~\cite{Hausler_2021_CVPR, 9699393,tolias2015particular} widely adopted. Most VG methods employ \textit{metric learning} with triplet, pair, or multi-similarity~\cite{multisim} loss for distinguishing positive samples from the same place and negative samples from different places. A recent benchmark~\cite{Berton_CVPR_2022_benchmark} highlights the effectiveness of triplet loss with Hard Negative Mining (HNM). Additionally, a trend toward classification-based VG methods~\cite{Berton_CVPR_2022_CosPlace, Trivigno_2023_ICCV, Berton_2023_EigenPlaces} is emerging, grouping images by location and distinguishing between different classes.

VG methods can also be classified into one-stage approaches~\cite{Berton_CVPR_2022_benchmark, Berton_CVPR_2022_CosPlace, netvlad, 8382272, ge2020self, Leyva-Vallina_2023_CVPR}, which use global image embeddings, and two-stage approaches~\cite{transvpr, r2former, Hausler_2021_CVPR, Sarlin_2020_CVPR}, which enhance accuracy by generating local features and re-ranking top candidates. While two-stage methods offer better performance, they require more memory and longer inference times. Compared with these works, our work focuses on benchmarking SSL approaches using geo-referenced imagery for geo-specific representation learning, establishing an SSL baseline compared to state-of-the-art VG methods.

{\bf Self-supervised Representation Learning.} Recent advances in self-supervised representation learning have achieved performance on par with, or even surpassing, supervised methods in computer vision tasks. These methods learn robust visual representations from unlabeled image data through advanced data augmentation. The key challenge in SSL is avoiding model collapse, where the model produces identical outputs regardless of input, making it difficult to differentiate samples. We focus on three categories of SSL methods: {\it Contrastive Learning}~\cite{simclr,moco,mocov2}, {\it Self-distillation Learning}~\cite{byol,SimSiam}, and {\it Information Maximization}~\cite{bt,vicreg}. A detailed architecture comparison is provided in Sec.\ref{cat} of the appendix.

\textit{Contrastive Learning.} Methods like SimCLR~\cite{simclr} and MoCov2~\cite{mocov2} align embeddings of augmented views of the same image while maximizing the embedding distance from unrelated images. To prevent collapse, SimCLR uses a large batch size and a projection head, while MoCo~\cite{moco} and MoCov2 introduce InfoNCE~\cite{CPC} loss and a momentum encoder, maintaining a queue of embeddings to avoid collapse without requiring large batch sizes.

\textit{Self-distillation Learning.} Methods like BYOL~\cite{byol} and SimSiam~\cite{SimSiam} prevent collapse using a teacher-student framework. BYOL uses a momentum target encoder and cosine distance loss between predicted and actual embeddings. SimSiam highlights the importance of stop-gradient operations and batch normalization in avoiding collapse.

\textit{Information maximization.} Methods like Barlow Twins \cite{bt} and VICReg~\cite{vicreg} decorrelate embeddings to maximize information. Barlow Twins uses a cross-correlation matrix to suppress off-diagonal elements, while VICReg employs Variance-Invariance-Covariance (VIC) Regularization to maintain variance and reduce covariance between embedding elements.

\textbf{Intersection between VG and SSL.} Recent works have explored the intersection of VG and SSL methods. For instance, \cite{remotesensorssl} reviews SSL techniques for remote sensing classification within a pretrain-finetune framework. \cite{ayush2021geography} employs the MoCov2~\cite{mocov2} framework with a memory dictionary, using geographical information prediction as a pretext task. \cite{ali2023global} improves hard negative mining in VG through a MoCo-like memory bank. Unlike these studies, our work focuses on a benchmarking framework for SSL methods in the VG context, introducing the GeoPair strategy. GeoPair links VG metric learning with SSL by using geo-labels to capture real geo-relationships and employing SSL techniques to enhance sample efficiency by data augmentations.

 \begin{figure*}[h]
    \centering
    \includegraphics[width=\textwidth]{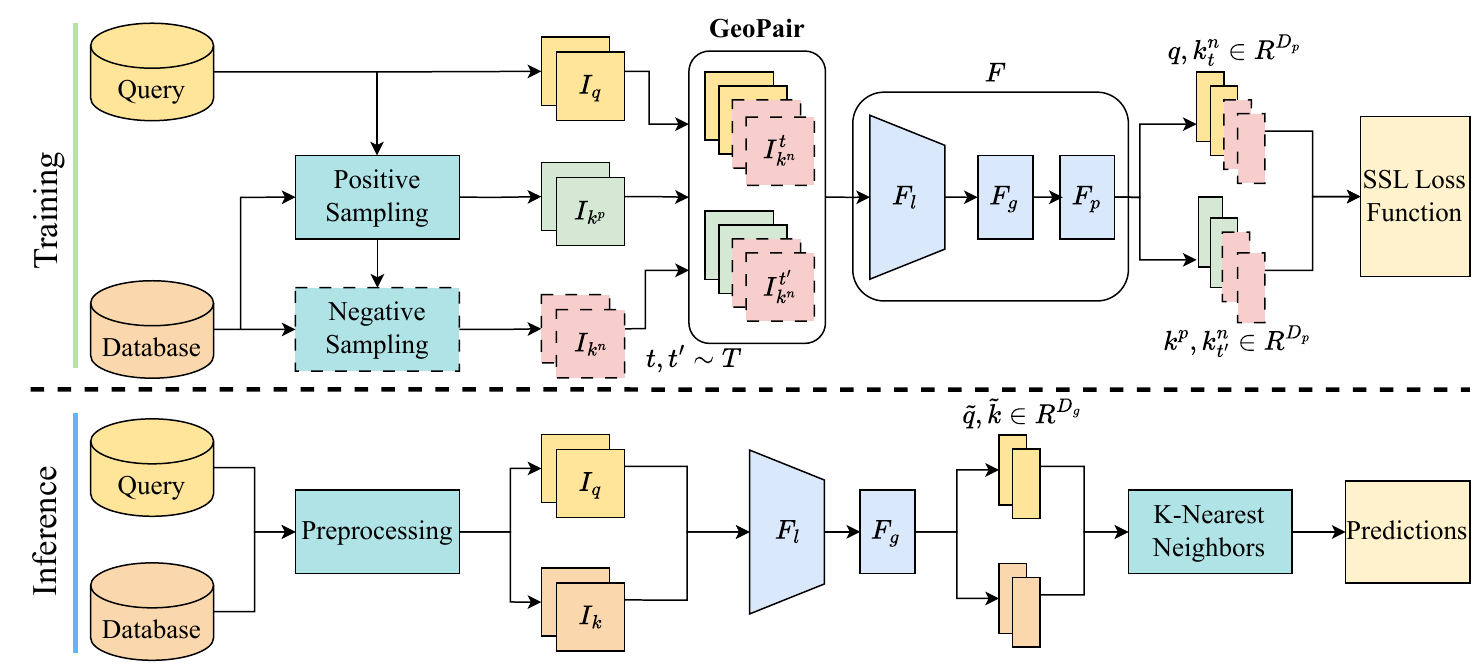}
    \caption{VG-SSL Architecture: During training, query images $I_q$ and positive database images $I_{k^p}$ are sampled, with optional negative images $I_{k^n}$ selected via HNM. \textbf{GeoPair} strategy builds image pairs using query-positive pairs $I_q$, $I_{k^p}$ and augmented negative pairs $I^{t}_{k^n}$, $I^{t^\prime}_{k^n}$ with augmentation $t, t^\prime \sim T$. The feature extractor $F$ then produces embeddings ($q$, $k^p$, $k_{t}^n$, and $k_{t^\prime}^n$), and SSL loss is applied to train $F$. During inference, the projection head is removed, and KNN is used with feature embeddings $\tilde q$ and $\tilde k$ from the aggregation module.}
     \label{fig:framework}
\end{figure*}
\section{Methodology}\label{sec:methodology}

In our VG-SSL framework, depicted in Fig.~\ref{fig:framework}, we have designed the GeoPair strategy to build image pairs to accommodate multiple SSL training strategies. This strategy provides query images ($I_q$), positive images ($I_{k^p}$), and augmented negative images ($I^{t}_{k^n}$ and $I^{t^\prime}_{k^n}$) for the feature extractor $F$. The extractor comprises a local feature extractor $F_l$ (e.g., ResNet-50~\cite{resnet}), a global feature aggregation module $F_g$ (e.g., GeM~\cite{8382272}), and a projection head $F_p$ (an MLP with consistent hidden and output dimensionality), inspired by \cite{simclr}, to map feature embeddings into a projected embedding space. The projection head is a common component in SSL methods for enhancing embedding quality. Subsequently, we design a unified module that allows the selection of various SSL loss functions to utilize these projected embeddings for training $F$.


We denote $D_g$ as the dimension of feature embeddings output by $F_g$, $D_p$ as the dimension of projected embeddings output by $F_p$, $N$ as the batch size, $L$ as the number of layers in the projection head, $q$ as the projected embeddings of $I_q$, $k^p$ and $k^n$ as the projected embeddings of $I_{k^p}$ and $I_{k^n}$, respectively. The terms $q_b = \begin{bmatrix}
q_{b,1} &  q_{b,2}& \cdots &  q_{b,D_p}\end{bmatrix}^\top$ and $k^p_{b} = \begin{bmatrix} k^p_{b,1} & k^p_{b,2} & \cdots & k^p_{b,D_p}\end{bmatrix}^\top$ are the $b$-th embedding with dimension $D_p$ of queries and positives respectively. The terms $q=\begin{bmatrix}q_1 & q_2 & \cdots  & q_N\end{bmatrix}$ and $k^p = \begin{bmatrix}k^p_1 & k^p_2 & \cdots & k^p_N\end{bmatrix}$ are the batch of paired embeddings for training. Optionally, if we choose to use negative samples in training, we employ Hard Negative Mining (HNM) to select negatives. Our preliminary results show that the model trained with randomly selected negatives yields performance similar to that of those trained with only query-positive pairs. We use data augmentation $t, t^\prime \sim T$ to augment $I_{k^n}$ and get augmented views $I^{t}_{k^n}$ and $I^{t^\prime}_{k^n}$. We concatenate the negatives with queries and positives and build the batch of paired embeddings as $q^\prime = \begin{bmatrix}q_1 & q_2 & ...  & q_N & k^n_{1,t} &k^n_{2,t} &... & k^n_{N,t}\end{bmatrix}$ and ${k^p}^\prime = \begin{bmatrix}k^p_1 & k^p_2 & \cdots  & k^p_N & k^n_{1,t^\prime} & k^n_{2,t^\prime} & \cdots  & k^n_{N,t^\prime}\end{bmatrix}$.

At the inference stage, the preprocessed query and database images input the feature extractor. The preprocessing is detailed in Sec.~\ref{preprocess} of the appendix. $\Tilde{q}$ and $\Tilde{k}$ denote the feature embeddings of $I_q$ and $I_k$, respectively. By computing $L_2$ distance between query and database feature embeddings, we use the K-Nearest Neighbors (KNN) algorithm to identify the top-K candidates.

\textbf{GeoPair:} The key idea of the GeoPair strategy is to merge metric learning from VG with the SSL paradigm. Metric learning leverages Hard Negative Mining (HNM) to form triplets or pairs that distinguish between positive ($I_{k^p}$), negative ($I_{k^n}$), and query samples ($I_{q}$). SSL, on the other hand, uses data augmentation to create varied views ($I^{t}_{k^n}$ and $I^\prime_{k^n}$) of a single instance as positives, treats other instances as negatives, and employs advanced loss functions to enhance sampling efficiency. GeoPair builds geo-specific query-positive pairs using geo-labels and incorporates augmented views of negatives, facilitating the learning of real geographic relationships and robust representation learning following the SSL paradigm.
 
\subsection{Loss functions}

The VG-SSL framework integrates selected SSL loss functions, highlighting a critical distinction between metric learning and SSL in terms of the embeddings targeted by the loss function. Specifically, metric learning directly optimizes the distance metrics of feature embeddings produced by 
$F_g$. On the other hand, SSL approaches optimize projected embeddings generated by $F_p$ to be invariant to data augmentation. Subsequently, during the inference phase, SSL methods revert to utilizing the original, unprojected feature embeddings output by $F_g$.
 




{\bf InfoNCE Loss.} InfoNCE Loss \cite{CPC} is used in contrastive learning SSL methods \cite{simclr, moco, mocov2}. The intuition of this loss is to reduce the embedding distance between positive pairs and increase the embedding distance to all other pairs. The loss function is formulated as
\begin{equation}\label{infonce}\small\mathcal{L}_{\textrm{InfoNCE}} = \frac{1}{N}\sum_{b=1}^N\log\frac{\exp(s(q_b, k^p_{b})/\tau)}{\sum_{i=1, i\neq b}^{N}\exp(s(q_b, k^p_{i})/\tau)},
\end{equation}
where $\tau$ is the temperature parameter, and $s(q_b, k^p_{b}) = \frac{q_b\cdot k^p_{b}}{\vert\vert q_b\vert\vert_2 \vert\vert  k^p_{b}\vert\vert_2}$.

{\bf Embedding Prediction Loss.} Embedding Prediction Loss is used in self-distillation methods \cite{byol, SimSiam}. The loss function aims to predict the common representation between the teacher embedding $k^p_b$ and the student embedding $q_b$. The loss function is formulated as
\begin{equation}
\small
    \mathcal{L}_{\textrm{EmbPred}} = \frac{1}{N}\sum_{b=1}^N[2 - 2\frac{p(q_b)\cdot\textrm{stopgrad}(k^p_{b})}{\vert\vert p(q_b)\vert\vert_2\vert\vert\textrm{stopgrad}(k^p_{b})\vert\vert_2}],
\end{equation}
where $p(\cdot)$ is a shallow MLP to predict $k^p_b$ from $q_b$ and $\textrm{stopgrad}(\cdot)$ means the gradient is not backpropagated to the teacher model.

{\bf Cross-correlation Loss.} Barlow Twins (BT) \cite{bt} introduces the Cross-correlation Loss. First, it builds a cross-correlation matrix defined as
\begin{equation}
\small
    C_{ij} = \frac{\sum_{b=1}^Nq_{b,i}k^p_{b,j}}{\sqrt{\sum_{b=1}^N(q_{b,i})^2}\sqrt{\sum_{b=1}^N(k^p_{b,j})^2}},
\end{equation}
where $i,j$ are the index of elements of the embedding, and $C$ is a square cross-correlation matrix range from -1 (negative correlation) to 1 (positive correlation). Then, the BT loss $L_\textrm{BT}$ is calculated as
\begin{equation}\small
    \mathcal{L}_\textrm{BT} = \sum_{i=1}^D (1 - C_{ii})^2 + \lambda \sum_{i=1}^D \sum_{j=1,j\neq i}^D C_{ij},
\end{equation}
where $\lambda$ is the weight of the off-diagonal term as a hyperparameter. The intuition of this loss is to enforce the same element of embeddings positively correlated and the different elements of embeddings uncorrelated to maximize contained information.

{\bf VIC Regularization Loss.} VIC Regularization (VICReg) Loss \cite{vicreg} is inspired by BT Loss and explicitly defines the invariance (first row), variance (second row), and covariance (third row) terms of the elements of the embeddings as
\begin{equation}
\small
\begin{split}
     \mathcal{L}_\textrm{VICReg} &= \frac{\lambda_1}{ND}\sum_{b=1}^N\sum_{i=1}^D \Vert q_{b,i} - k^p_{b,i} \Vert_2^2\\
     & + \frac{\lambda_2}{D}\sum_{i=1}^D[\max\{d_2 - \textrm{std}_i(q), 0\} + \max\{d_2 - \textrm{std}_i(k^{p}), 0\}]\\
     & + \frac{\lambda_3}{D}\sum_{i=1}^D\sum_{j=1,j\neq i}^D\left [\textrm{cov}^2_{ij}(q) + \textrm{cov}^2_{ij}(k^p) \right],\\
\end{split}
\end{equation}
where $\lambda_1, \lambda_2, \lambda_3, d_2$ are scalars, $\textrm{std}_i(q)$ is the standard deviation of the $i$-th element of $q$ over the batch, and $\textrm{cov}_{ij}(q)$ is the element of the $i$-th row and $j$-th column of the covariance matrix of $q$ over the batch. This loss function decorrelates elements of the embedding by reducing the off-diagonal elements of the covariance matrix to zeros. It also maximizes the standard deviation of the element of the embeddings over the batch with a margin $d_2$.

\section{Experimental Setup}\label{sec:experiement}

\subsection{Datasets and Metrics}
We employ five VG datasets featuring handhold and car-mounted camera settings that are widely used in robotics and autonomous driving applications: Pitts30k~\cite{pitts30k}, MSLS~\cite{msls}, Tokyo24/7~\cite{tokyo247}, Nordland~\cite{Sunderhauf_2013_nordland}, and Revisited San Francisco (R-SF)~\cite{Chen_2011_san_francisco, RSF} (Reported in Appendix Sec.~\ref{san}). We leverage the expansive MSLS dataset for both training and evaluation. For evaluation, the Pitts30k and Tokyo24/7 datasets are primarily composed of urban imagery, with both their database images derived from Google Street View. The query set of Tokyo24/7 is distinct, consisting of images captured via smartphone. The Nordland dataset presents a similar viewpoint to MSLS and R-SF but specifically introduces challenges related to seasonal change from summer to winter.

Our experimental evaluation is anchored by the Recall@N (R@N) metric, which assesses the proportion of query images that are correctly identified within the top-N predictions, provided that they fall within a pre-set proximity of $25$ meters from the true geo-location.  This threshold aligns with prior research in the VG domain for a fair comparison. This metric is a pivotal indicator of VG accuracy, with our analysis focusing on R@1, R@5, and R@10.

\subsection{Implementation Details}
In our study, we adopt certain protocols from the benchmark VG pipeline as outlined in \cite{Berton_CVPR_2022_benchmark}, while incorporating GeoPair to accommodate various SSL techniques within the VG framework. The training image size is $480\times 640$ ($H\times W$). With a batch size $B$ of $64$, each batch consists of $64$ image pairs. We use Adam optimizer \cite{kingma2014adam} with learning rates $1e^{-5}$. We instantiate both one-stage and two-stage methods with different SSL methods. For one-stage methods, we conduct our experiments mainly with ResNet-50~\cite{resnet} + GeM~\cite{8382272} or DeiT-S~\cite{touvron2021training} (Distilled version of ViT-S~\cite{dosovitskiy2020vit}) used in \cite{Berton_CVPR_2022_benchmark,r2former}. For two-stage methods, we primarily use DeiT-S for the first stage and the reranking part of R2Former~\cite{r2former} for the second stage. We run each experiment for $36$ hours with approximately $200$ epochs. The data augmentation is set to be random resizing and cropping with a random scale from 1.0 to 1.25, and random flipping ($p=0.5$). During training, we define images as positive samples if they are within the $10$ m radius of the query images and negative samples if they are beyond the $25$ m radius of the query images. The models are primarily trained with a batch size of $64$ (ResNet-50 + GeM) or $32$ (DeiT-S) in one NVIDIA-A100-80GB GPU, while the final model is trained with a batch size of $32$, $64$, $128$, or $256$, depending on the final performance. The pipelines are implemented using Pytorch-lightning \cite{Falcon_PyTorch_Lightning_2019}. 

\textbf{Remarks on Data Augmentation.} As noted in \cite{Berton_CVPR_2022_benchmark}, the effectiveness of data augmentation can vary across datasets due to their unique characteristics. To establish a robust baseline, we fully utilize standard data augmentations, including random flipping and random resized cropping, which avoid dataset-specific biases and provide general benefits across all datasets. While we agree that exploring more advanced augmentations could enhance performance, applying them across multiple datasets can be complex and may result in inconsistent outcomes. We suggest investigating the impact of specific augmentations when fine-tuning models on individual datasets for more targeted improvements.
\section{Results}\label{sec:results}
\subsection{Comparison with State-of-the-art}
Table \ref{sota} presents a comparison between the best results of various SSL methods and state-of-the-art one-stage VG methods, including NetVLAD~\cite{netvlad}, SFRS~\cite{ge2020self}, GCL-GeM~\cite{Leyva-Vallina_2023_CVPR}, TransVPR~\cite{transvpr} (without re-ranking), and R2Former~\cite{r2former} (without re-ranking) for both ResNet-50~\cite{resnet} and ResNeXt~\cite{Xie_2017_CVPR}. To ensure a fair comparison, all methods were trained on the MSLS and Pitts30k datasets. Details about the state-of-the-art results can be found in Sec.\ref{source} of the appendix. Due to page limitations, extended results for two-stage methods, including SP-SuperGlue\cite{Sarlin_2020_CVPR}, Patch-NetVLAD~\cite{Hausler_2021_CVPR}, TransVPR~\cite{transvpr}, and R2Former~\cite{r2former}, are provided in Sec.~\ref{twostage} of the appendix.

\textbf{Inference Complexity:} For our one-stage methods, we use ResNet-50~\cite{resnet} + GeM~\cite{8382272} with $D_g=1024$ and DeiT-S~\cite{touvron2021training} with $D_g=256$. This ensures that memory usage and inference time are consistent with ResNet50-GeM~\cite{Berton_CVPR_2022_benchmark} and R2Former~\cite{r2former} without re-ranking.

\definecolor{dgreen}{rgb}{0.60, 0.87, 0.48}
\definecolor{yellow}{rgb}{0.93, 0.95, 0.59}
\definecolor{dred}{rgb}{1.0, 0.56, 0.56}

\begin{table*}[ht]
\centering
\caption{Comparison of state-of-the-art VG methods with our results on large-scale VG datasets. Our models were initially trained on the MSLS dataset. For evaluating performance in urban environments (Pitts30k and Tokyo24/7), we further fine-tuned our models on the Pitts30k dataset. The best results are highlighted in \textbf{bold}, with colors indicating performance tiers of our results: \textcolor{dgreen}{green} for the top 3 results, \textcolor{yellow}{yellow} for the next 2 results, and \textcolor{dred}{red} for the 2 lowest-performing results. $^*$ denotes the performance without reranking.}\label{sota}
\scriptsize
\resizebox{\linewidth}{!}{
\begin{tabular}{lccccccccccccccccccccc}
\toprule
    &\multirow{2}{1em}{$D_g$}&
    \multicolumn{3}{c}{MSLS Val} & & \multicolumn{3}{c}{MSLS Challenge} & & \multicolumn{3}{c}{Pitts30k} & & \multicolumn{3}{c}{Tokyo24/7} & & \multicolumn{3}{c}{Nordland}\\
    \cline{3-5}\cline{7-9}\cline{11-13}\cline{15-17}\cline{19-21}
    & & R@1 & R@5 &  R@10&  & R@1 & R@5 &  R@10& & R@1 & R@5 &  R@10& & R@1 & R@5 &  R@10& & R@1 & R@5 &  R@10\\\midrule
    \multicolumn{10}{l}{\textit{One-Stage Methods}} \\
    NetVLAD~\cite{netvlad} & - & 60.8 & 74.3 & 79.5 & &35.1 & 47.4 & 51.7& &81.9 &  91.2 & 93.7 & &64.8 &  78.4& 81.6 & & - & - & -   \\
    SFRS~\cite{ge2020self} & 4096&  69.2& 80.3 & 83.1& & 41.5& 52.0& 56.3& & \textbf{89.4}& \textbf{94.7}& 95.9& &  \textbf{85.4} &  \textbf{91.1} & \textbf{93.3} & & 18.8 & 32.8 & 39.8 \\
    TransVPR$^*$~\cite{transvpr} &  256 &70.8 & 85.1 &  89.6 & & 48.0 &  67.1 & 73.6 & & 73.8 &  88.1 & 91.9& & - & - &- & & 15.9 & 38.6 & 49.4\\
    R2Former$^*$~\cite{r2former} &  256 & 79.3 & 90.5 & 92.7 & & 54.9 & 75.1 & 79.6  & &72.9 & 88.5 & 92.6 & & 43.5 & 65.7 & 72.4 & &21.4 & 33.7 & 41.0  \\
    GCL-ResNet50-GeM~\cite{Leyva-Vallina_2023_CVPR} & 1024 & 74.6 &  84.7&  88.1 & &  52.9&  65.7&  71.9 & & 79.9 &   90.0& 92.8 & &  58.7& 71.1& 76.8 & & - & - & -    \\
    GCL-ResNeXt-GeM~\cite{Leyva-Vallina_2023_CVPR} &  1024 &80.9 & 90.7& 92.6 & & 62.3& 76.2& 81.1 & & 79.2 &  90.4& 93.2 & &  58.1& 74.3& 78.1 & & - & - & -    \\
    \midrule
    \multicolumn{10}{l}{\textit{Our One-Stage Methods with ResNet50-GeM}} \\
    Triplet Loss (Baseline)~\cite{Berton_CVPR_2022_CosPlace} & 1024 & \cellcolor{yellow}{76.9} & \cellcolor{yellow}{86.1} & \cellcolor{yellow}{89.5} & & \cellcolor{yellow}{53.5} & \cellcolor{yellow}{68.1} & \cellcolor{yellow}{72.3} & & \cellcolor{dred}{76.7} & \cellcolor{dred}{89.1} & \cellcolor{dred}{92.3} & & \cellcolor{yellow}{50.2} & \cellcolor{dgreen}{67.9} & \cellcolor{dgreen}{76.8} & & \cellcolor{dgreen}{39.5} & \cellcolor{dgreen}{\textbf{59.0}} & \cellcolor{dgreen}{\textbf{67.7}}\\
    SimCLR  & 1024 & \cellcolor{dgreen}{\textbf{84.2}} & \cellcolor{dgreen}{\textbf{92.2}} & \cellcolor{dgreen}{\textbf{94.2}} & & \cellcolor{dgreen}{\textbf{63.1}} & \cellcolor{dgreen}{\textbf{78.9}} & \cellcolor{dgreen}{\textbf{83.6}} & & \cellcolor{dgreen}{82.8} & \cellcolor{dgreen}{91.9}&  \cellcolor{dgreen}{94.6} & & \cellcolor{dgreen}{54.6} & \cellcolor{dgreen}{74.9} & \cellcolor{dgreen}{81.9}& & \cellcolor{dgreen}{\textbf{39.9}} & \cellcolor{dgreen}{56.4} & \cellcolor{dgreen}{63.9} \\
    MoCov2 & 1024 & \cellcolor{dgreen}{81.5} & \cellcolor{dgreen}{90.5} & \cellcolor{dgreen}{92.8} & & \cellcolor{dgreen}{59.0} & \cellcolor{yellow}{73.8} & \cellcolor{dgreen}{79.2}& & \cellcolor{dgreen}{82.6} & \cellcolor{dgreen}{92.4}&  \cellcolor{dgreen}{95.1} & & \cellcolor{dgreen}{51.4} & \cellcolor{dgreen}{68.3} & \cellcolor{dgreen}{76.5} & & \cellcolor{dgreen}{28.0}& \cellcolor{dgreen}{42.7} & \cellcolor{dgreen}{50.1} \\
    BYOL & 1024 & \cellcolor{dred}{72.7} & \cellcolor{dred}{85.5} & \cellcolor{dred}{87.7} & & \cellcolor{dred}{50.4} & \cellcolor{dred}{66.4} & \cellcolor{dred}{71.4} & & \cellcolor{yellow}{80.2} &\cellcolor{yellow}{91.5} & \cellcolor{dgreen}{94.4} & &\cellcolor{dred}{44.8} &\cellcolor{dred}{63.8} &\cellcolor{dred}{70.8} & &  \cellcolor{dred}{10.6}& \cellcolor{dred}{18.5} & \cellcolor{dred}{23.5} \\
    SimSiam & 1024 & \cellcolor{dred}{75.0} & \cellcolor{dred}{85.8} & \cellcolor{dred}{88.6} & & \cellcolor{dred}{52.1} & \cellcolor{dred}{67.0} & \cellcolor{dred}{72.2} & & \cellcolor{dred}{78.6} & \cellcolor{dred}{89.8} & \cellcolor{dred}{92.7} & &\cellcolor{dgreen}{51.1} &\cellcolor{yellow}{67.6} &\cellcolor{yellow}{71.4} & &  \cellcolor{dred}{12.5}& \cellcolor{dred}{21.5} & \cellcolor{dred}{27.0}  \\
    Barlow Twins & 1024 & \cellcolor{dgreen}{79.5} & \cellcolor{dgreen}{89.5} & \cellcolor{dgreen}{91.9} & & \cellcolor{dgreen}{59.2} &\cellcolor{dgreen}{74.2} & \cellcolor{dgreen}{79.1} & & \cellcolor{dgreen}{80.8} & \cellcolor{dgreen}{91.7} & \cellcolor{yellow}{94.2} & &\cellcolor{dred}{45.7} &\cellcolor{dred}{61.9} &\cellcolor{dred}{70.8} & &  \cellcolor{yellow}{18.5}& \cellcolor{yellow}{30.5} & \cellcolor{yellow}{38.0}  \\
    VICReg & 1024 & \cellcolor{yellow}{77.4} & \cellcolor{yellow}{89.3} & \cellcolor{yellow}{91.2} & & \cellcolor{yellow}{58.0} & \cellcolor{dgreen}{74.1} & \cellcolor{yellow}{79.0} & & \cellcolor{yellow}{80.2} & \cellcolor{yellow}{91.3} & \cellcolor{yellow}{94.1} & &\cellcolor{yellow}{50.2} & \cellcolor{yellow}{65.4} &\cellcolor{yellow}{74.3} & &  \cellcolor{yellow}{14.9}& \cellcolor{yellow}{25.1} & \cellcolor{yellow}{31.3}  \\\midrule
    \multicolumn{10}{l}{\textit{Our One-Stage Methods with DeiT-S}} \\
    Triplet Loss (Baseline)~\cite{r2former} &  256 & \cellcolor{dgreen}{79.3} & \cellcolor{dgreen}{90.5} & \cellcolor{dgreen}{92.7} & & \cellcolor{yellow}{54.9} & \cellcolor{yellow}{75.1} & \cellcolor{dgreen}{79.6}  & &\cellcolor{dred}{72.9} & \cellcolor{dred}{88.5} & \cellcolor{dred}{92.6} & & \cellcolor{dred}{43.5} & \cellcolor{yellow}{65.7} & \cellcolor{dred}{72.4} & &\cellcolor{dgreen}{21.4} & \cellcolor{dgreen}{33.7} & \cellcolor{dgreen}{41.0}  \\
    SimCLR & 256 & \cellcolor{dgreen}{81.1} & \cellcolor{dgreen}{91.1} & \cellcolor{dgreen}{93.1} & &\cellcolor{dgreen}{58.9} & \cellcolor{dgreen}{77.1} & \cellcolor{dgreen}{82.6}& &\cellcolor{dgreen}{84.7} &\cellcolor{dgreen}{93.9} &\cellcolor{dgreen}{\textbf{96.0}} & &\cellcolor{dgreen}{59.4} &\cellcolor{dgreen}{76.2} &\cellcolor{dgreen}{80.0} & &\cellcolor{dgreen}{24.9} & \cellcolor{dgreen}{38.9} & \cellcolor{dgreen}{46.1}\\
    MoCov2 & 256 & \cellcolor{yellow}{76.1} & \cellcolor{yellow}{88.5} & \cellcolor{yellow}{91.1}& & \cellcolor{yellow}{56.8} & \cellcolor{dgreen}{75.2} & \cellcolor{yellow}{78.7}& &\cellcolor{yellow}{80.8} &\cellcolor{dgreen}{92.4} &\cellcolor{yellow}{95.0} & &\cellcolor{yellow}{50.8} &\cellcolor{dgreen}{69.8} &\cellcolor{dgreen}{77.1} & & \cellcolor{yellow}{15.4} & \cellcolor{yellow}{26.4} & \cellcolor{yellow}{33.0} \\
    BYOL & 256 & \cellcolor{dred}{58.2} & \cellcolor{dred}{75.3} & \cellcolor{dred}{79.6} & & \cellcolor{dred}{37.7} & \cellcolor{dred}{54.0} & \cellcolor{dred}{60.4} & &\cellcolor{dred}{76.6} &\cellcolor{dred}{89.4}&\cellcolor{dred}{92.9} & &\cellcolor{dred}{43.2} &\cellcolor{dred}{62.2} &\cellcolor{dred}{68.6} & &\cellcolor{dred}{4.1} & \cellcolor{dred}{7.9} & \cellcolor{dred}{10.6}\\
    SimSiam & 256 & \cellcolor{dred}{56.2} & \cellcolor{dred}{76.2}& \cellcolor{dred}{80.1} & & \cellcolor{dred}{35.3} & \cellcolor{dred}{52.3} & \cellcolor{dred}{58.7} & &\cellcolor{yellow}{79.7} &\cellcolor{yellow}{91.0} &\cellcolor{yellow}{93.6} & &\cellcolor{yellow}{47.3} &\cellcolor{dred}{63.8} &\cellcolor{yellow}{74.0} & & \cellcolor{dred}{6.2} & \cellcolor{dred}{11.5} & \cellcolor{dred}{15.4} \\
    Barlow Twins & 256 & \cellcolor{dgreen}{79.7} & \cellcolor{dgreen}{91.4} & \cellcolor{dgreen}{93.1}& & \cellcolor{dgreen}{59.1} & \cellcolor{dgreen}{76.1} & \cellcolor{dgreen}{81.5}& &\cellcolor{dgreen}{82.6} &\cellcolor{yellow}{92.1} &\cellcolor{dgreen}{95.0} & &\cellcolor{dgreen}{58.4} &\cellcolor{dgreen}{75.2} &\cellcolor{dgreen}{80.6} & & \cellcolor{dgreen}{28.1} & \cellcolor{dgreen}{43.3} & \cellcolor{dgreen}{51.1} \\
    VICReg & 256 & \cellcolor{yellow}{75.8} & \cellcolor{dgreen}{89.5} & \cellcolor{yellow}{91.9} & & \cellcolor{dgreen}{56.9} & \cellcolor{yellow}{74.0} & \cellcolor{yellow}{78.2}& &\cellcolor{dgreen}{81.7} &\cellcolor{dgreen}{92.3} &\cellcolor{dgreen}{95.2} & &\cellcolor{dgreen}{51.7} &\cellcolor{yellow}{66.7} &\cellcolor{yellow}{74.6} & &\cellcolor{yellow}{19.3}  &\cellcolor{yellow}{32.1} &\cellcolor{yellow}{39.6}   \\
    \bottomrule
\end{tabular}
}
\end{table*}

In Table~\ref{sota}, our comparison of different Self-Supervised Learning (SSL) strategies shows that SimCLR, MoCov2, and Barlow Twins consistently perform well across various datasets, including MSLS (Val and Challenge subsets), Pitts30k, and Nordland, demonstrating their versatility and effectiveness in diverse environments, from general landscapes to urban settings. SimCLR stands out for its consistently strong performance across all datasets, highlighting the robustness of contrastive learning. These results emphasize the power of both contrastive learning and information maximization strategies.

When comparing ResNet50-GeM ($D_g=1024$) and DeiT-S ($D_g=256$) variants, we observed a drop in performance for SimCLR, MoCov2, BYOL, and SimSiam on MSLS and Nordland, particularly affecting the R@1 metric in contrastive learning methods and overall recall in self-distillation methods. This indicates the significant impact of embedding dimension ($D_g$) on performance. In contrast, information maximization methods like Barlow Twins and VICReg maintained or improved their metrics with a smaller $D_g$, suggesting greater robustness to reduced dimensions.

Compared to state-of-the-art one-stage methods, SimCLR, MoCov2, and Barlow Twins exhibit competitive or superior performance on the MSLS and Nordland datasets, likely due to their advanced feature learning capabilities. SFRS~\cite{ge2020self}, while excelling in urban datasets like Pitts30k and Tokyo24/7, relies on a much larger embedding dimension ($D_g=4096$), highlighting a trade-off between computational complexity and performance. However, SFRS shows limited generalization in more diverse environments like MSLS and Nordland, possibly due to its training bias toward urban landscapes.

On the other hand, GCL methods~\cite{Leyva-Vallina_2023_CVPR} perform similarly to our approaches with $D_g=1024$, but they require additional processing steps like Principal Component Analysis (PCA) and orientation label training. In contrast, our SSL-based methods avoid such extra steps, using smaller embedding dimensions ($D_g=1024$ for ResNet50-GeM and $D_g=256$ for DeiT-S), reducing computational overhead and improving inference efficiency. This underscores not only the competitiveness of our SSL strategies with state-of-the-art methods but also the practical efficiency of our approach.

\subsection{Ablation study}
In this section, we conduct an ablation study with our one-stage methods on the following key factors influencing VG performance: Hard Negative Mining (HNM), Number of Projection Layers $L$, Dimensionality of Projected Embeddings $D_p$ and Feature Embeddings $D_g$. We report MSLS Val's performance for the study. If not specifically mentioned, we set the following parameters from the initial hyperparameter search: $F$ is ResNet50-GeM; Training batch size is $64$; SimCLR, BT, and VICReg use HNM; MoCov2, BYOL, and SimSiam do not use HNM; $D_p=2048$; $D_g=1024$; SimCLR, MoCov2 use $L=1$; BYOL, SimSiam, BT, VICReg use $L=2$.

\begin{table*}[ht]
\centering
\caption{Comparison of training with hard negative mining, different number of projection layers $L$, different dimension of projected embeddings $D_p$, and different dimension of feature embeddings $D_g$. The best results for each setting are highlighted in \textbf{bold}. \xmark \, means that the model cannot converge. - means the model is skipped since the performance has saturated.}\label{ab}
\resizebox{0.9\linewidth}{!}{
\begin{tabular}{cccccccccccccccccccccccc}
\toprule
     & \multicolumn{3}{c}{SimCLR} & & \multicolumn{3}{c}{MoCov2} & &\multicolumn{3}{c}{BYOL} & & \multicolumn{3}{c}{SimSiam}& &\multicolumn{3}{c}{Barlow Twins}& & \multicolumn{3}{c}{VICReg}\\
     \cline{2-4}\cline{6-8}\cline{10-12}\cline{14-16}\cline{18-20}\cline{22-24}
     & R@1 & R@5 &  R@10&  & R@1 & R@5 &  R@10& & R@1 & R@5 &  R@10& & R@1 & R@5 & R@10 & &R@1 & R@5 &  R@10 & &R@1 & R@5 &  R@10\\\midrule
     \multicolumn{10}{l}{\textit{Hard Negative Mining}} \\
     & 79.6 & 90.8 & 92.0 & & \textbf{77.8} & \textbf{89.3} & \textbf{91.9} & &\textbf{73.0} & 83.2 & 85.9 & &\textbf{74.3} & \textbf{84.9} & \textbf{87.4} & &65.5 & 82.7 & 85.8 & &66.1 & 81.4 & 85.3\\
    \cmark & \textbf{82.4} & \textbf{91.4} & \textbf{93.4} & &76.5 & 86.4 & 88.2 & &71.1 & \textbf{83.9} & \textbf{86.2} & &71.8 & 83.0 & 85.7 & &\textbf{79.5} & \textbf{89.5} & \textbf{91.9} & &\textbf{77.6} & \textbf{86.9} & \textbf{89.2} \\\midrule
    \multicolumn{10}{l}{\textit{Number of Projection Layers $L$}} \\
    0 & 80.9 & \textbf{91.9} &93.1 & &\textbf{80.5} & \textbf{90.1} & \textbf{91.9}& & \xmark & \xmark & \xmark & &\xmark & \xmark & \xmark & & 76.6 & 86.6 & 88.5 & & \xmark & \xmark & \xmark\\
    1 &\textbf{82.4} & 91.4 & \textbf{93.4}& & 77.8 & 89.3 & \textbf{91.9} & & \xmark & \xmark & \xmark & &\xmark & \xmark & \xmark & & 78.6 & 87.6 & 89.5 & &\xmark & \xmark & \xmark\\
    2 & \textbf{82.4} & 90.1 & 91.8 & & 77.0 & 88.0 & 89.6 & & \textbf{73.0} & 83.2 & \textbf{85.9} & & \textbf{74.3} & \textbf{84.9} & \textbf{87.4} & & \textbf{79.5} & \textbf{89.5} & \textbf{91.9} & & \textbf{77.6} & 86.9 & 89.2\\
    3 & - & - & - & & - & - & - & & 70.4 & \textbf{83.8} & 85.5 & & 69.3 & 83.9 & 87.3 & &78.8 & 89.1 & 91.4 & & 76.9 & \textbf{87.8} & \textbf{90.0} \\\midrule
    \multicolumn{10}{l}{\textit{Dimension of projected embeddings $D_p$}} \\
    1024 & \textbf{84.2} & \textbf{92.2} & \textbf{94.3} & & 78.5 & 89.1 & \textbf{93.2} & & 71.5 & 83.9 & 86.9 & & 72.4 & 84.5 & 87.0 & &76.4 & 87.6 & 89.9  & &74.1 & 86.1 & 88.2\\
    2048 & 82.4 & 91.4 & 93.4 & &  77.8 & \textbf{89.3} & 91.9& &\textbf{73.0} & 83.2 & 85.9& & 74.3 & 84.9 & 87.4& & \textbf{79.5} & \textbf{89.5} & \textbf{91.9}& &77.6 & 86.9 & 89.2\\
    4096 & 82.3 & 91.8 & 93.0& &\textbf{79.9} & \textbf{89.3} & 91.4& &72.8 & \textbf{85.5} & \textbf{87.7} & &\textbf{75.0} & \textbf{85.8} & \textbf{88.6}& &76.1 & 87.7 & 90.4& &\textbf{77.7} & \textbf{88.8} & \textbf{90.7}\\
    \midrule
    \multicolumn{10}{l}{\textit{Dimension of feature embeddings $D_g$}}\\
    512 & 81.6 & \textbf{91.5} & \textbf{93.8} & & \textbf{79.5} & 89.2 & 90.8 & & 71.9 & 84.9 & 87.6 & & 72.8  & 86.4 & 88.6& & 76.4 & 87.8  & 90.7 & &77.3 & \textbf{87.7} & 89.7 \\
    1024 & 82.4 & 91.4 & 93.4 & & 77.8 & \textbf{89.3} & \textbf{91.9}& &73.0 & 83.2 & 85.9 & &74.3 & 84.9 & 87.4& &\textbf{79.5} & \textbf{89.5} & \textbf{91.9}& & 77.6 & 86.9 & 89.2\\
    2048 & \textbf{83.0} & 91.2 & 92.6  & & 78.4 & 89.1 & 91.4 & &\textbf{74.6} & \textbf{85.8} & \textbf{88.4}& &\textbf{74.6} & \textbf{86.4} & \textbf{88.8}& &79.2 & 89.3 & 91.4& &\textbf{78.5} & \textbf{87.7} & \textbf{89.9}\\
    \bottomrule
\end{tabular}
}
\vspace{-10pt}
\end{table*}

\subsubsection{Hard Negative Mining}
In Table \ref{ab}, we present the results associated with training models with and without the implementation of HNM. It is evident that models such as SimCLR, BT, and VICReg demonstrate enhanced performance when HNM is applied. Conversely, MoCov2 and SimSiam exhibit superior performance in the absence of HNM. The results for BYOL exhibit variability in performance with and without HNM application. However, the notably higher R@1 without HNM suggests a slight preference for training without HNM.


\subsubsection{Number of Projection Layers}

In Table \ref{ab}, we analyze the impact of the number of projection layers ($L$) on SSL model convergence. Notably, methods like BYOL, SimSiam, and VICReg fail to converge with fewer than two layers, as a single layer only supports linear transformations, while two or more layers enable more complex, non-linear transformations that enhance the representational power of the embeddings. In contrast, SimCLR performs best with one layer, while MoCov2 excels with none. Both BYOL and SimSiam perform well at $L=2$, and BT shows improved performance with two layers. VICReg's performance varies between $L=2$ and $L=3$, indicating a more complex relationship between projection layers and VG performance.

\begin{figure*}[]
\smallskip
\smallskip
    \centering
\rotatebox{90}{\scriptsize \phantom{H}}
\begin{subfigure}[b]{0.11\textwidth}
\centering
\tiny
Input
\end{subfigure}
\begin{subfigure}[b]{0.11\textwidth}
\centering
\tiny
Triplet Loss
\end{subfigure}
\begin{subfigure}[b]{0.11\textwidth}
\centering
\tiny
SimCLR
\end{subfigure}
\begin{subfigure}[b]{0.11\textwidth}
\centering
\tiny
MoCov2
\end{subfigure}
\begin{subfigure}[b]{0.11\textwidth}
\centering
\tiny
BYOL
\end{subfigure}
\begin{subfigure}[b]{0.11\textwidth}
\centering
\tiny
SimSiam
\end{subfigure}
\begin{subfigure}[b]{0.11\textwidth}
\centering
\tiny
BT
\end{subfigure}
\begin{subfigure}[b]{0.11\textwidth}
\centering
\tiny
VICReg
\end{subfigure}

\rotatebox{90}{\scriptsize\hspace{-1.5em}MSLS}
\begin{subfigure}[b]{0.11\textwidth}
    \includegraphics[width=\textwidth,height=0.75\textwidth]{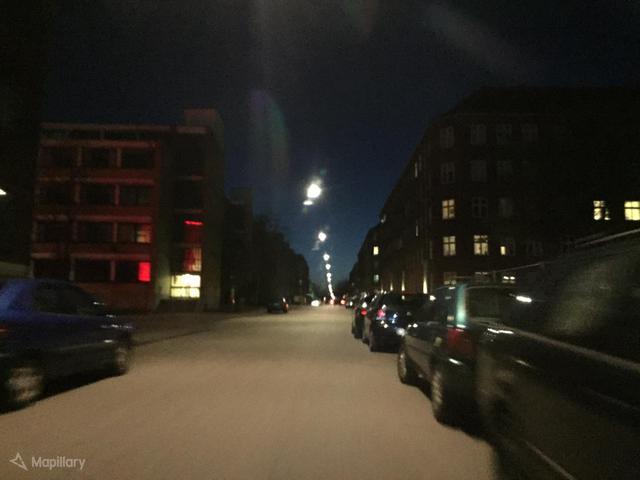}
    \vspace{-0.8\baselineskip}
\end{subfigure}
\begin{subfigure}[b]{0.11\textwidth}
    \includegraphics[width=\textwidth,height=0.75\textwidth]{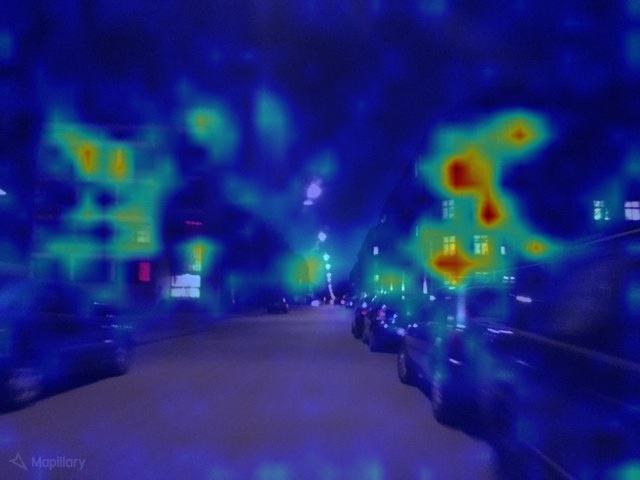}
    \vspace{-0.8\baselineskip}
\end{subfigure}
\begin{subfigure}[b]{0.11\textwidth}
    \includegraphics[width=\textwidth,height=0.75\textwidth]{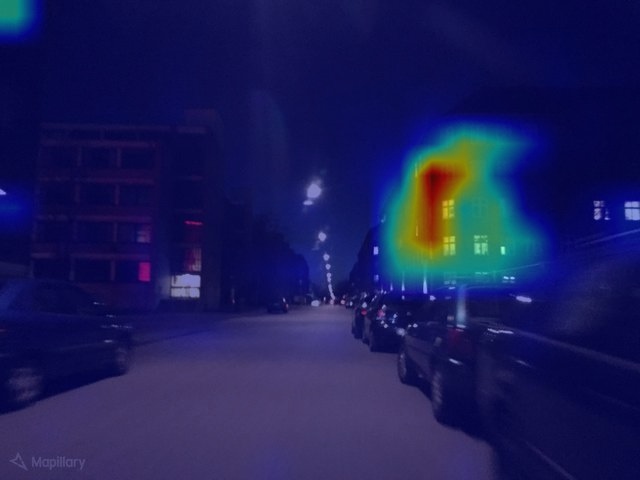}
    \vspace{-0.8\baselineskip}
\end{subfigure}
\begin{subfigure}[b]{0.11\textwidth}
    \includegraphics[width=\textwidth,height=0.75\textwidth]{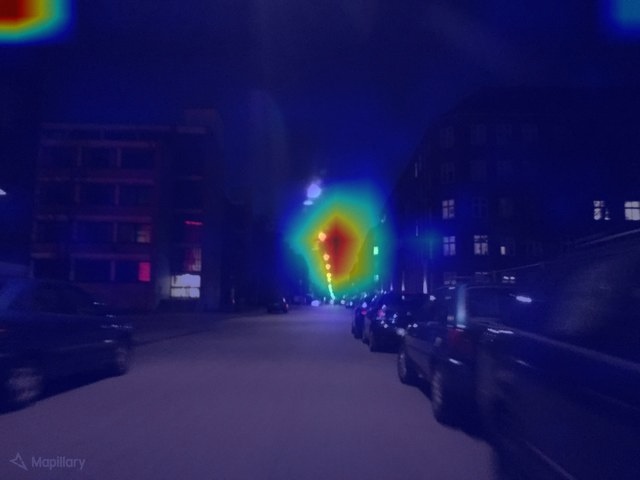}
    \vspace{-0.8\baselineskip}
\end{subfigure}
\begin{subfigure}[b]{0.11\textwidth}
    \includegraphics[width=\textwidth,height=0.75\textwidth]{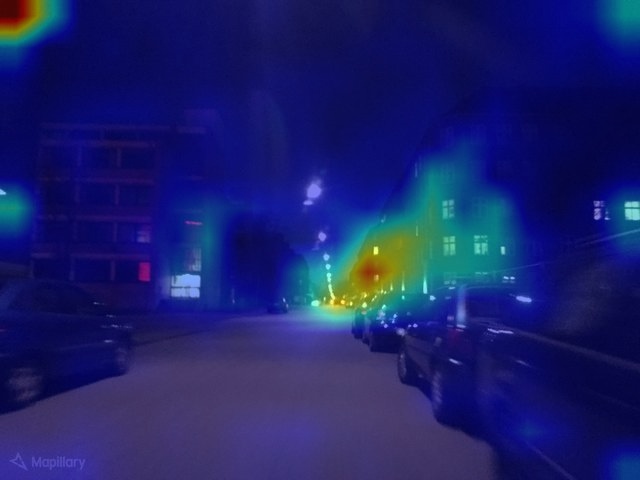}
    \vspace{-0.8\baselineskip}
\end{subfigure}
\begin{subfigure}[b]{0.11\textwidth}
    \includegraphics[width=\textwidth,height=0.75\textwidth]{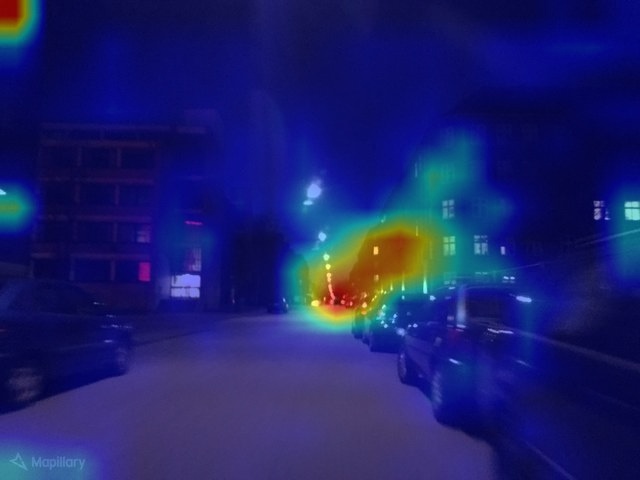}
    \vspace{-0.8\baselineskip}
\end{subfigure}
\begin{subfigure}[b]{0.11\textwidth}
    \includegraphics[width=\textwidth,height=0.75\textwidth]{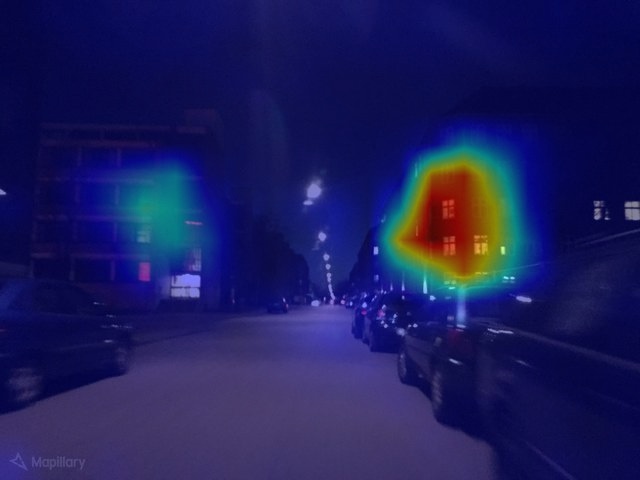}
    \vspace{-0.8\baselineskip}
\end{subfigure}
\begin{subfigure}[b]{0.11\textwidth}
    \includegraphics[width=\textwidth,height=0.75\textwidth]{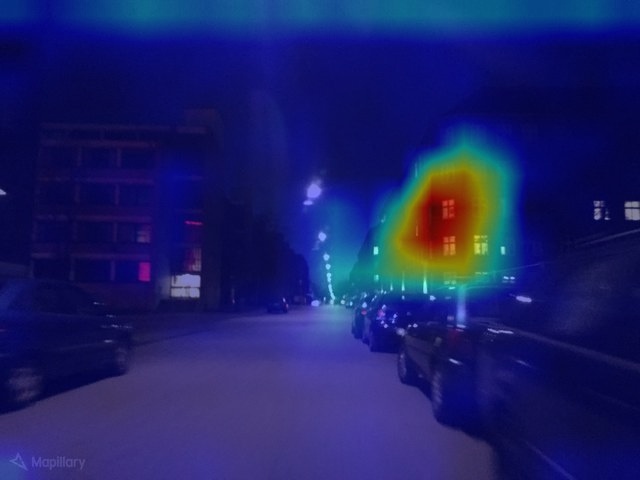}
    \vspace{-0.8\baselineskip}
\end{subfigure}

\rotatebox{90}{\scriptsize \phantom{H}}
\begin{subfigure}[b]{0.11\textwidth}
    \includegraphics[width=\textwidth,height=0.75\textwidth]{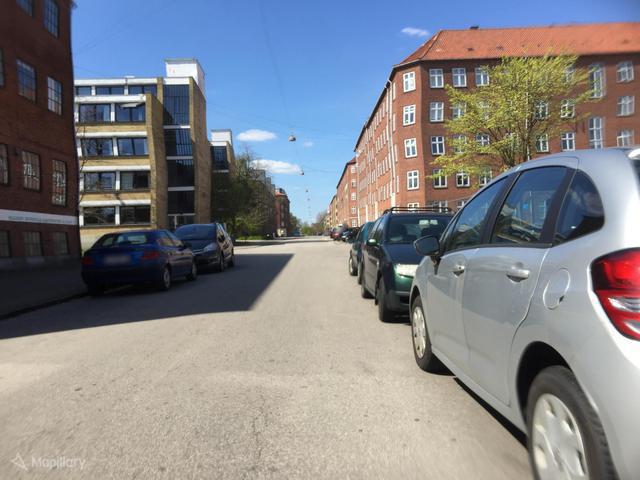}
    \hspace{5em}
    \vspace{-0.8\baselineskip}
\end{subfigure}
\begin{subfigure}[b]{0.11\textwidth}
    \includegraphics[width=\textwidth,height=0.75\textwidth]{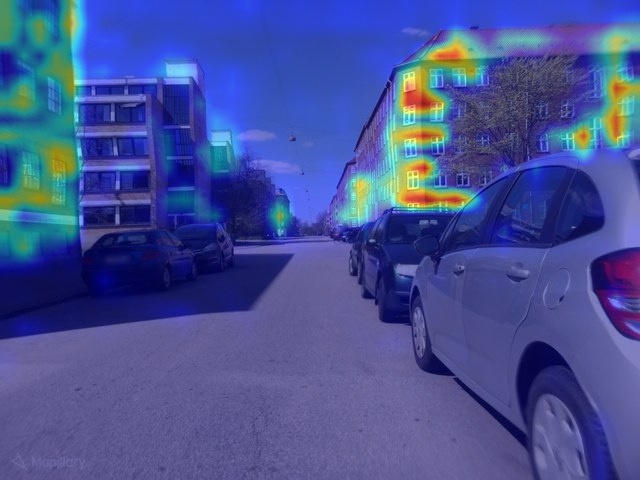}
    \vspace{-0.8\baselineskip}
\end{subfigure}
\begin{subfigure}[b]{0.11\textwidth}
    \includegraphics[width=\textwidth,height=0.75\textwidth]{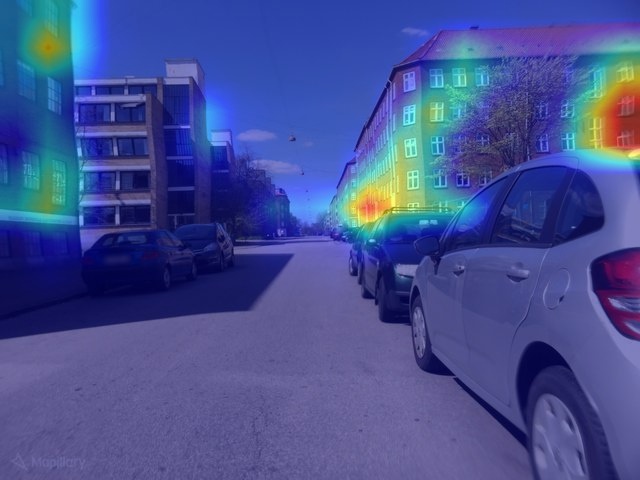}
    \vspace{-0.8\baselineskip}
\end{subfigure}
\begin{subfigure}[b]{0.11\textwidth}
    \includegraphics[width=\textwidth,height=0.75\textwidth]{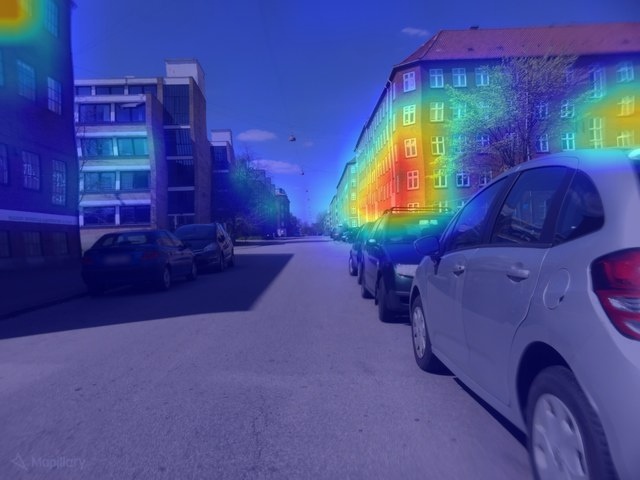}
    \vspace{-0.8\baselineskip}
\end{subfigure}
\begin{subfigure}[b]{0.11\textwidth}
    \includegraphics[width=\textwidth,height=0.75\textwidth]{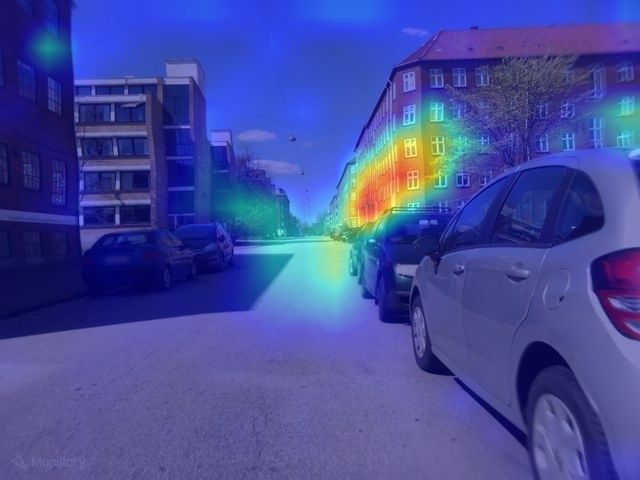}
    \vspace{-0.8\baselineskip}
\end{subfigure}
\begin{subfigure}[b]{0.11\textwidth}
    \includegraphics[width=\textwidth,height=0.75\textwidth]{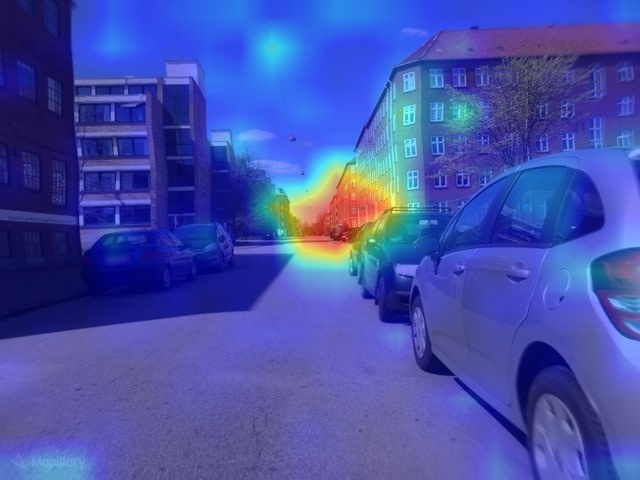}
    \vspace{-0.8\baselineskip}
\end{subfigure}
\begin{subfigure}[b]{0.11\textwidth}
    \includegraphics[width=\textwidth,height=0.75\textwidth]{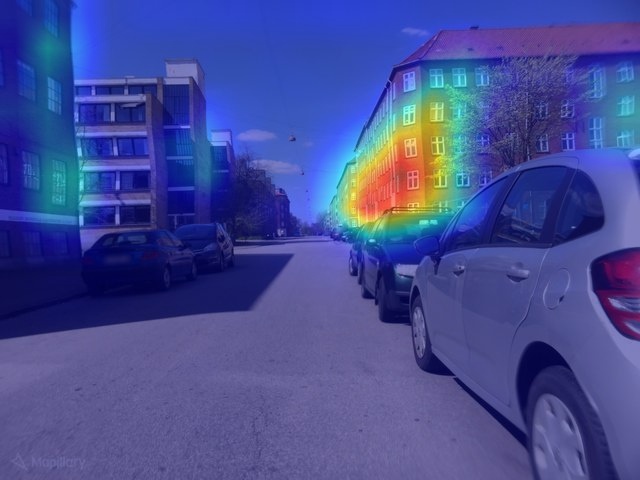}
    \vspace{-0.8\baselineskip}
\end{subfigure}
\begin{subfigure}[b]{0.11\textwidth}
    \includegraphics[width=\textwidth,height=0.75\textwidth]{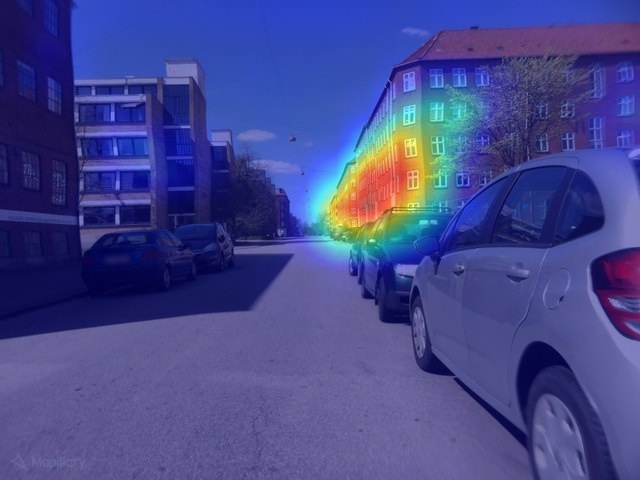}
    \vspace{-0.8\baselineskip}
\end{subfigure}

\rotatebox{90}{\scriptsize\hspace{-1.75em}Pitts30k}
\begin{subfigure}[b]{0.11\textwidth}
\includegraphics[width=\textwidth,height=0.75\textwidth]{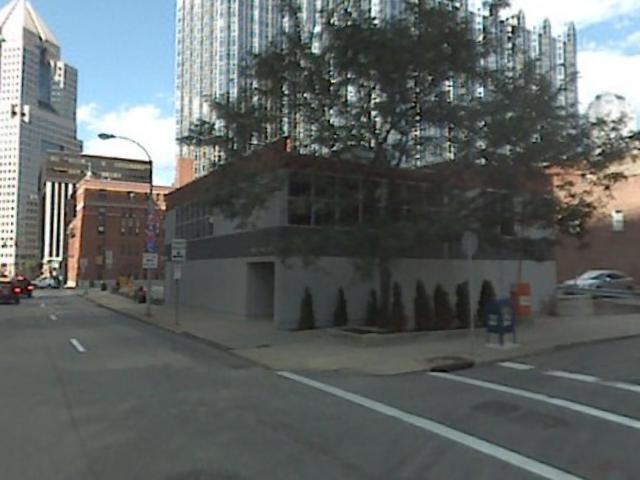}
    \vspace{-0.8\baselineskip}
\end{subfigure}
\begin{subfigure}[b]{0.11\textwidth}
\includegraphics[width=\textwidth,height=0.75\textwidth]{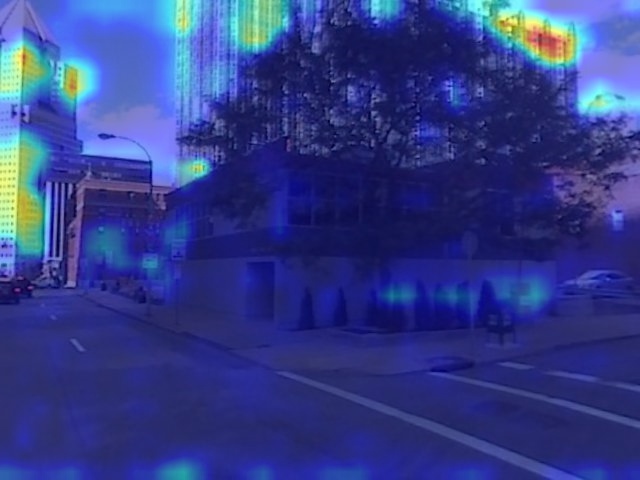}
    \vspace{-0.8\baselineskip}
\end{subfigure}
\begin{subfigure}[b]{0.11\textwidth}
\includegraphics[width=\textwidth,height=0.75\textwidth]{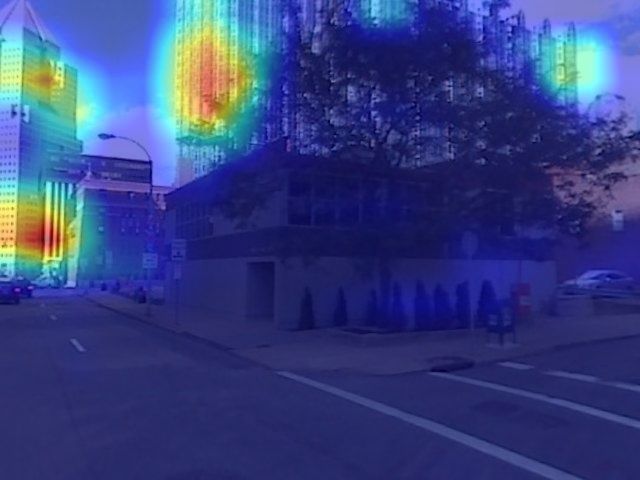}
    \vspace{-0.8\baselineskip}
\end{subfigure}
\begin{subfigure}[b]{0.11\textwidth}
\includegraphics[width=\textwidth,height=0.75\textwidth]{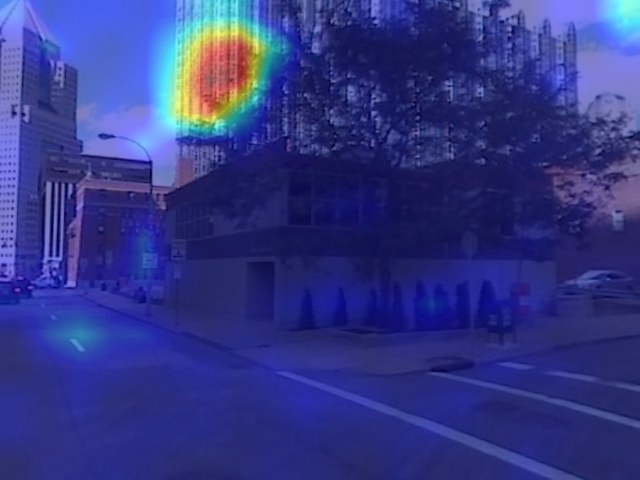}
    \vspace{-0.8\baselineskip}
\end{subfigure}
\begin{subfigure}[b]{0.11\textwidth}
\includegraphics[width=\textwidth,height=0.75\textwidth]{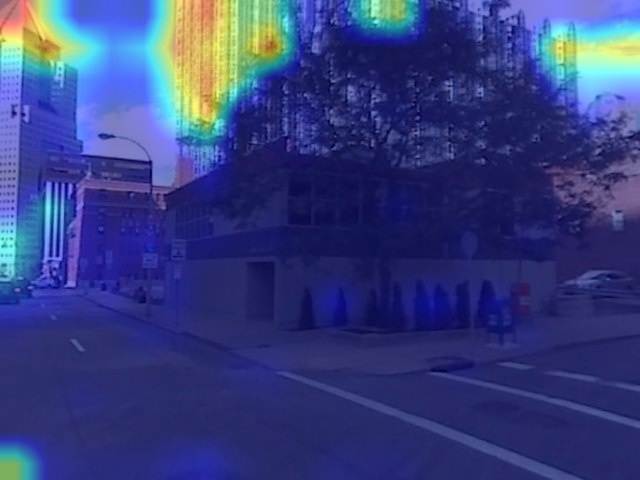}
    \vspace{-0.8\baselineskip}
\end{subfigure}
\begin{subfigure}[b]{0.11\textwidth}
\includegraphics[width=\textwidth,height=0.75\textwidth]{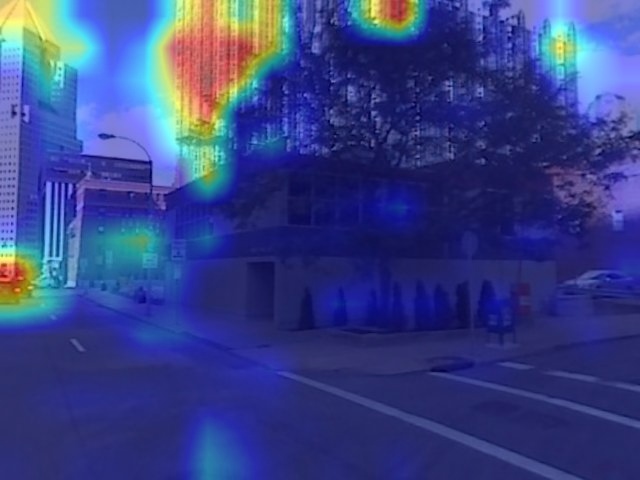}
    \vspace{-0.8\baselineskip}
\end{subfigure}
\begin{subfigure}[b]{0.11\textwidth}
\includegraphics[width=\textwidth,height=0.75\textwidth]{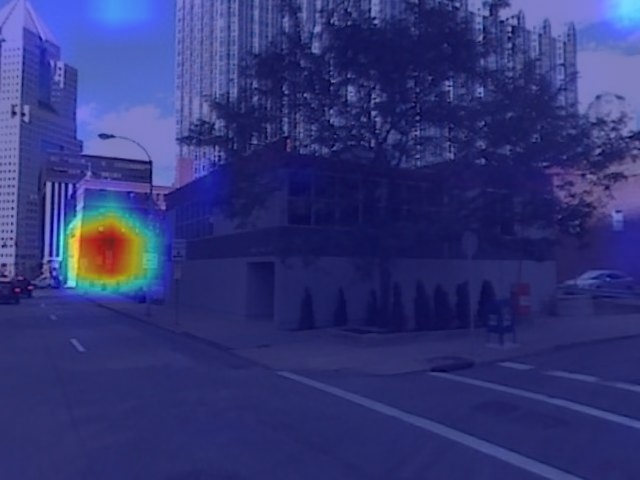}
    \vspace{-0.8\baselineskip}
\end{subfigure}
\begin{subfigure}[b]{0.11\textwidth}
\includegraphics[width=\textwidth,height=0.75\textwidth]{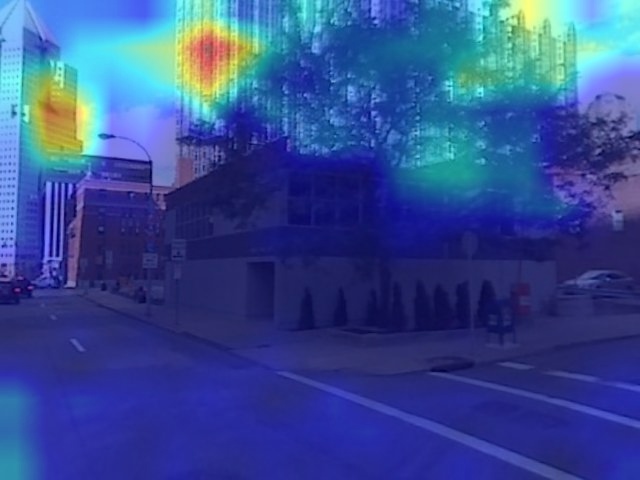}
    \vspace{-0.8\baselineskip}
\end{subfigure}

\rotatebox{90}{\scriptsize \phantom{H}}
\begin{subfigure}[b]{0.11\textwidth}
\includegraphics[width=\textwidth,height=0.75\textwidth]{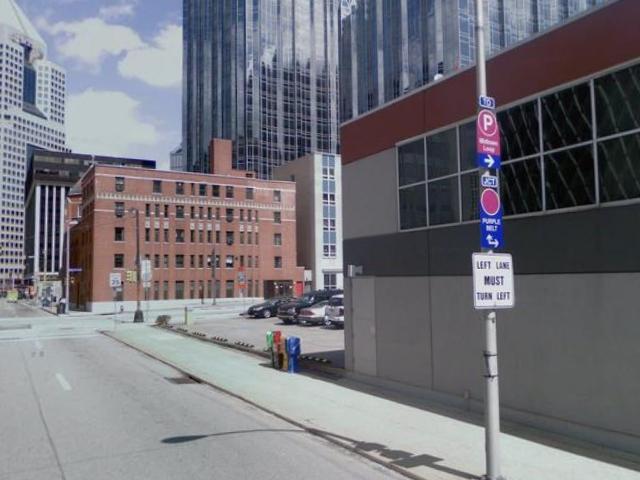}
    \vspace{-0.8\baselineskip}
\end{subfigure}
\begin{subfigure}[b]{0.11\textwidth}
\includegraphics[width=\textwidth,height=0.75\textwidth]{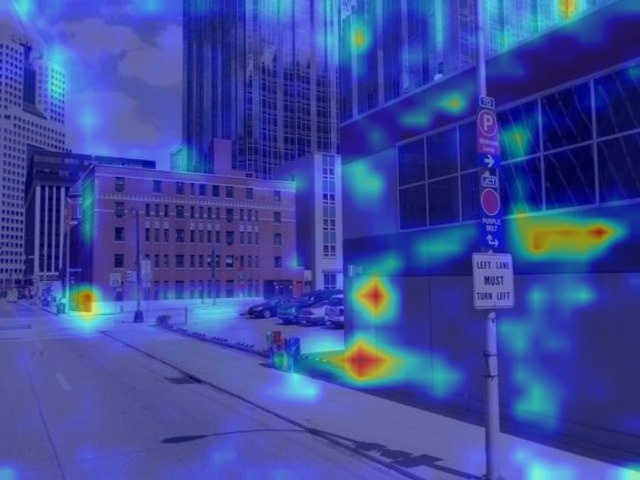}
    \vspace{-0.8\baselineskip}
\end{subfigure}
\begin{subfigure}[b]{0.11\textwidth}
\includegraphics[width=\textwidth,height=0.75\textwidth]{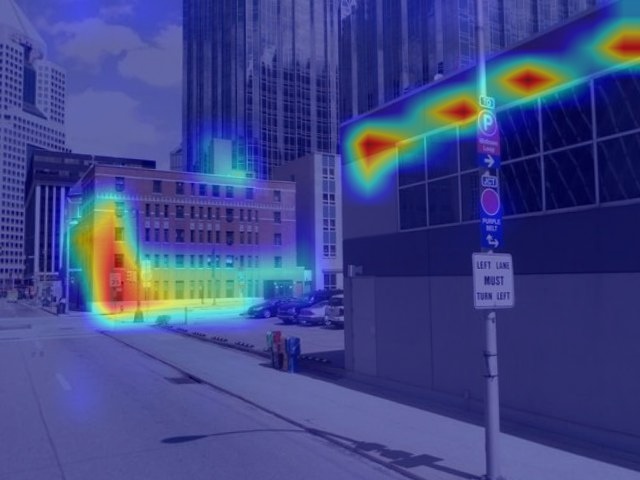}
    \vspace{-0.8\baselineskip}
\end{subfigure}
\begin{subfigure}[b]{0.11\textwidth}
\includegraphics[width=\textwidth,height=0.75\textwidth]{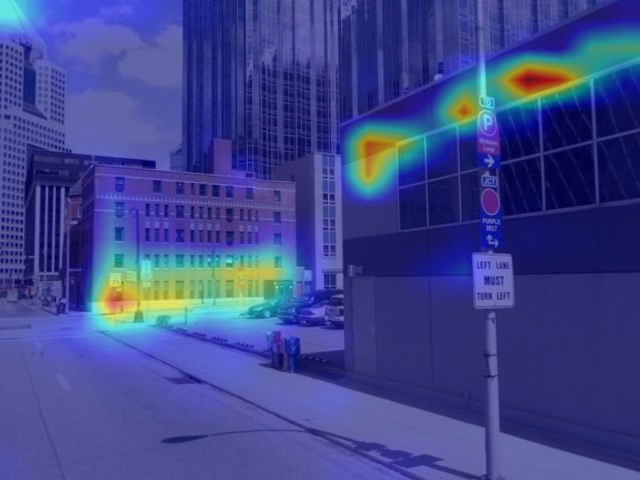}
    \vspace{-0.8\baselineskip}
\end{subfigure}
\begin{subfigure}[b]{0.11\textwidth}
\includegraphics[width=\textwidth,height=0.75\textwidth]{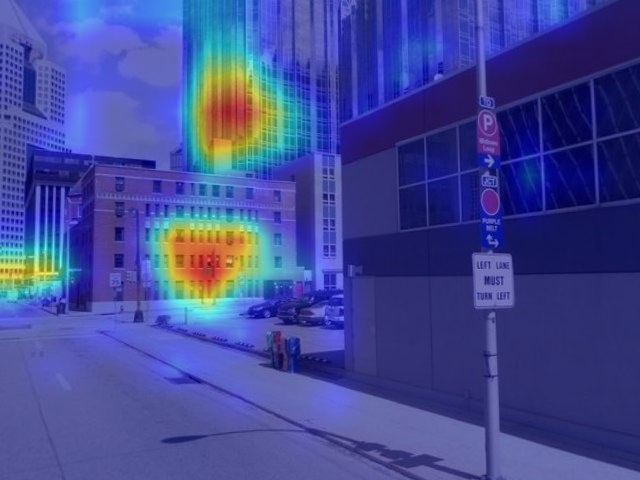}
    \vspace{-0.8\baselineskip}
\end{subfigure}
\begin{subfigure}[b]{0.11\textwidth}
\includegraphics[width=\textwidth,height=0.75\textwidth]{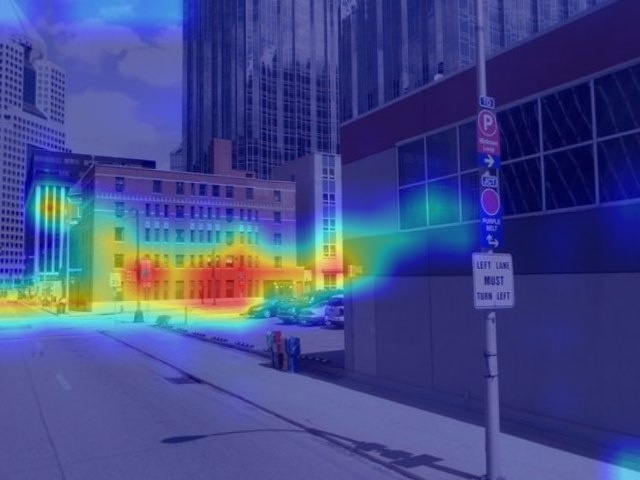}
    \vspace{-0.8\baselineskip}
\end{subfigure}
\begin{subfigure}[b]{0.11\textwidth}
\includegraphics[width=\textwidth,height=0.75\textwidth]{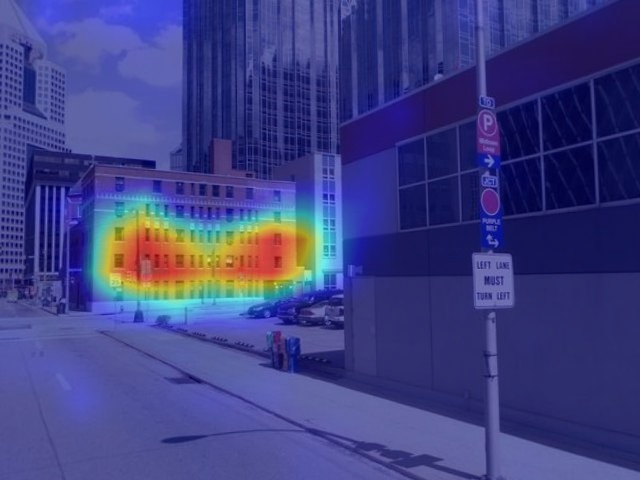}
    \vspace{-0.8\baselineskip}
\end{subfigure}
\begin{subfigure}[b]{0.11\textwidth}    \includegraphics[width=\textwidth,height=0.75\textwidth]{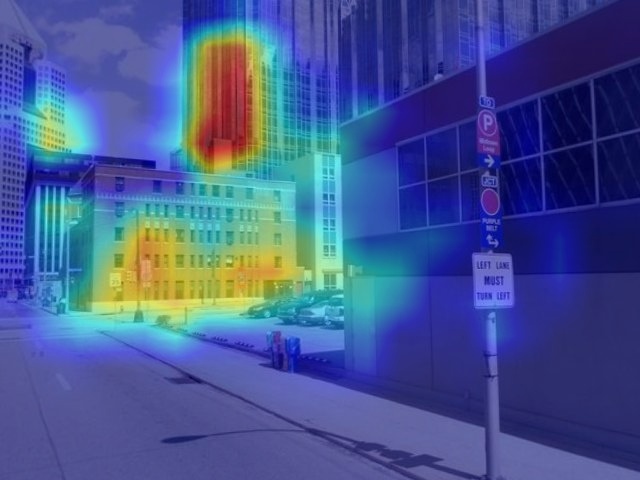}
    \vspace{-0.8\baselineskip}
\end{subfigure}

\rotatebox{90}{\tiny\hspace{-1.75em}Tokyo24/7}
\begin{subfigure}[b]{0.11\textwidth}
\includegraphics[width=\textwidth,height=0.75\textwidth]{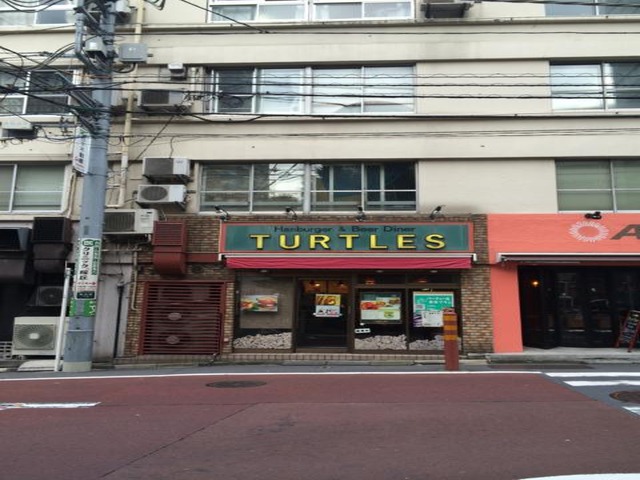}
    \vspace{-0.8\baselineskip}
\end{subfigure}
\begin{subfigure}[b]{0.11\textwidth}
\includegraphics[width=\textwidth,height=0.75\textwidth]{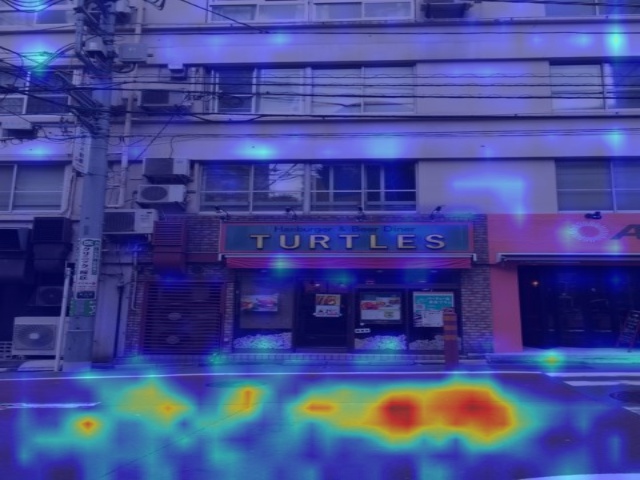}
    \vspace{-0.8\baselineskip}
\end{subfigure}
\begin{subfigure}[b]{0.11\textwidth}
\includegraphics[width=\textwidth,height=0.75\textwidth]{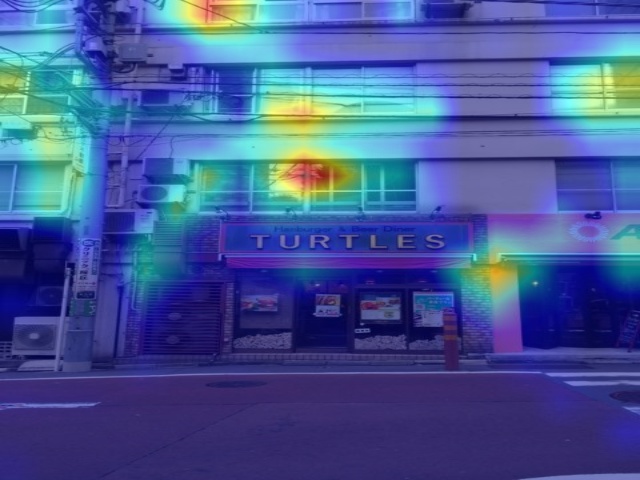}
    \vspace{-0.8\baselineskip}
\end{subfigure}
\begin{subfigure}[b]{0.11\textwidth}
\includegraphics[width=\textwidth,height=0.75\textwidth]{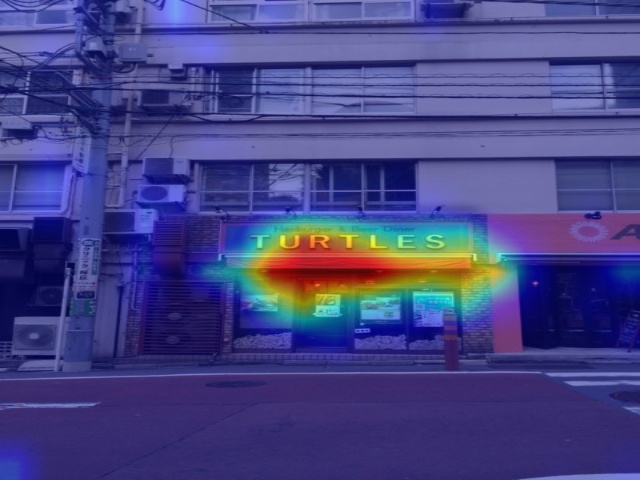}
    \vspace{-0.8\baselineskip}
\end{subfigure}
\begin{subfigure}[b]{0.11\textwidth}
\includegraphics[width=\textwidth,height=0.75\textwidth]{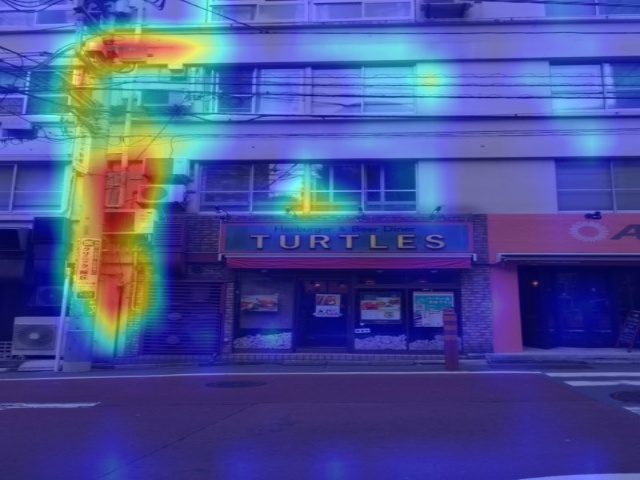}
    \vspace{-0.8\baselineskip}
\end{subfigure}
\begin{subfigure}[b]{0.11\textwidth}
\includegraphics[width=\textwidth,height=0.75\textwidth]{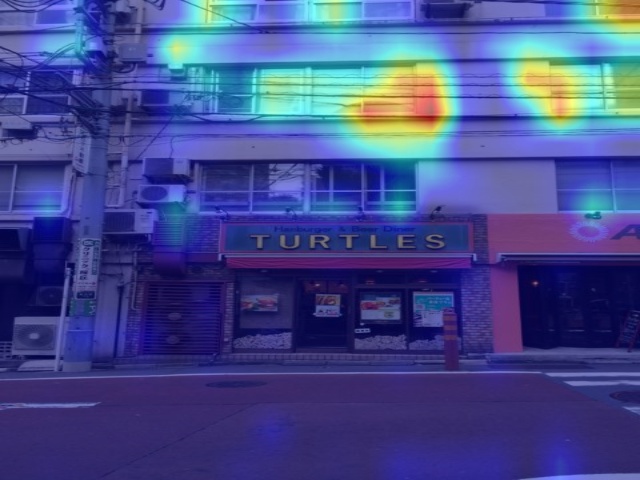}
    \vspace{-0.8\baselineskip}
\end{subfigure}
\begin{subfigure}[b]{0.11\textwidth}
\includegraphics[width=\textwidth,height=0.75\textwidth]{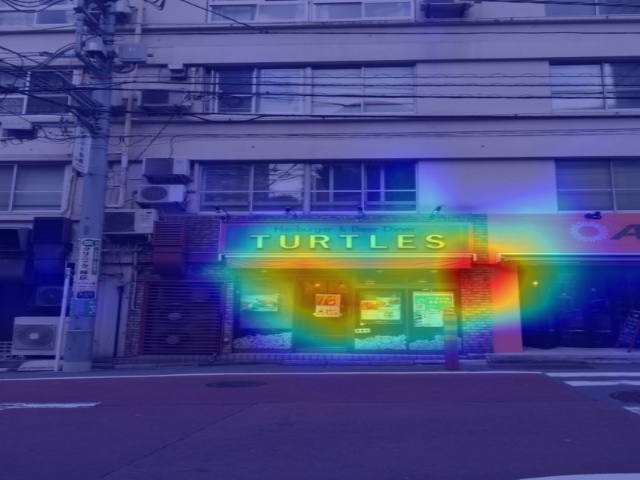}
    \vspace{-0.8\baselineskip}
\end{subfigure}
\begin{subfigure}[b]{0.11\textwidth}
\includegraphics[width=\textwidth,height=0.75\textwidth]{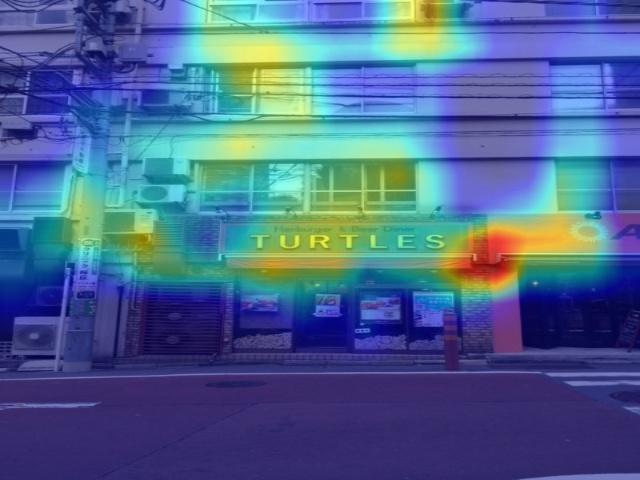}
    \vspace{-0.8\baselineskip}
\end{subfigure}

\rotatebox{90}{\scriptsize \phantom{H}}
\begin{subfigure}[b]{0.11\textwidth}
\includegraphics[width=\textwidth,height=0.75\textwidth]{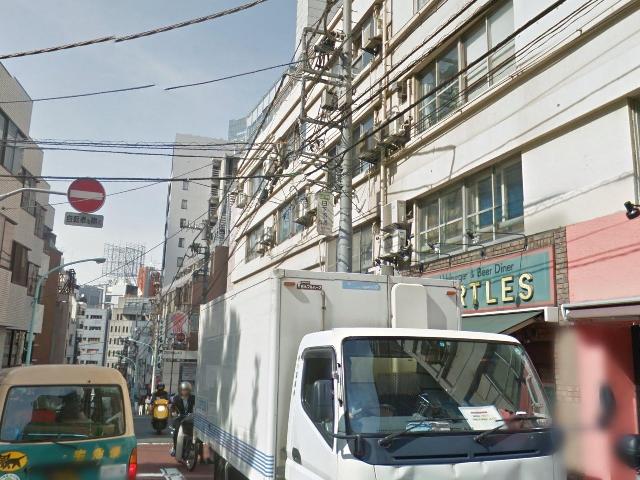}
    \vspace{-0.8\baselineskip}
\end{subfigure}
\begin{subfigure}[b]{0.11\textwidth}
\includegraphics[width=\textwidth,height=0.75\textwidth]{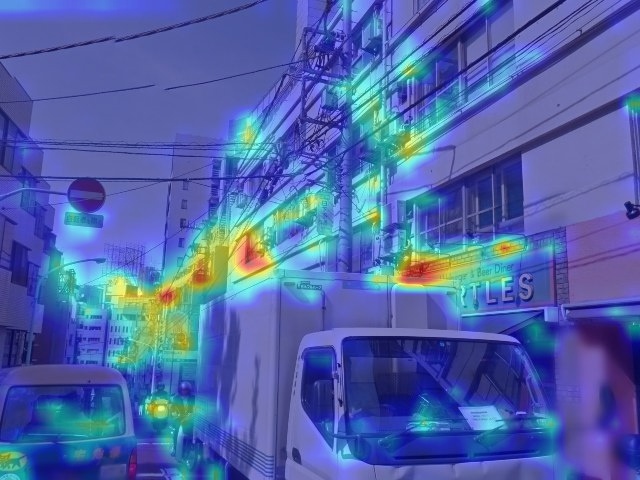}
    \vspace{-0.8\baselineskip}
\end{subfigure}
\begin{subfigure}[b]{0.11\textwidth}
\includegraphics[width=\textwidth,height=0.75\textwidth]{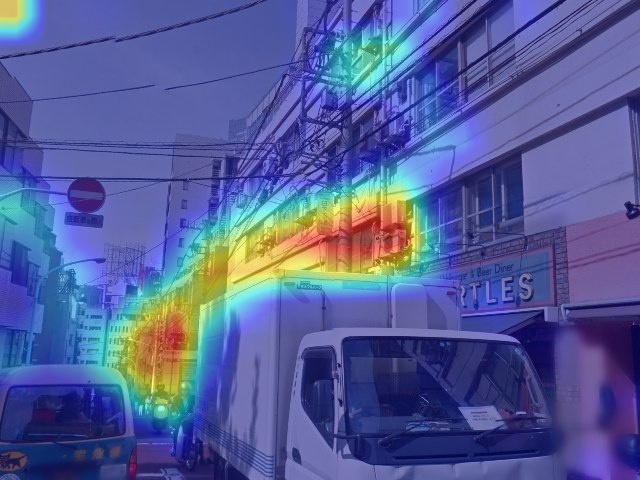}
    \vspace{-0.8\baselineskip}
\end{subfigure}
\begin{subfigure}[b]{0.11\textwidth}
\includegraphics[width=\textwidth,height=0.75\textwidth]{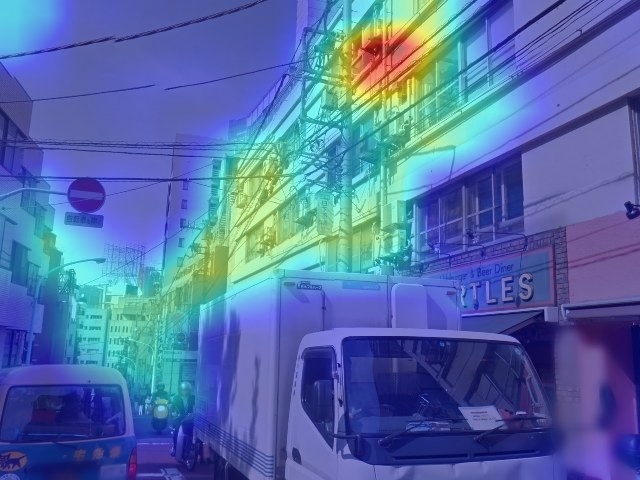}
    \vspace{-0.8\baselineskip}
\end{subfigure}
\begin{subfigure}[b]{0.11\textwidth}
\includegraphics[width=\textwidth,height=0.75\textwidth]{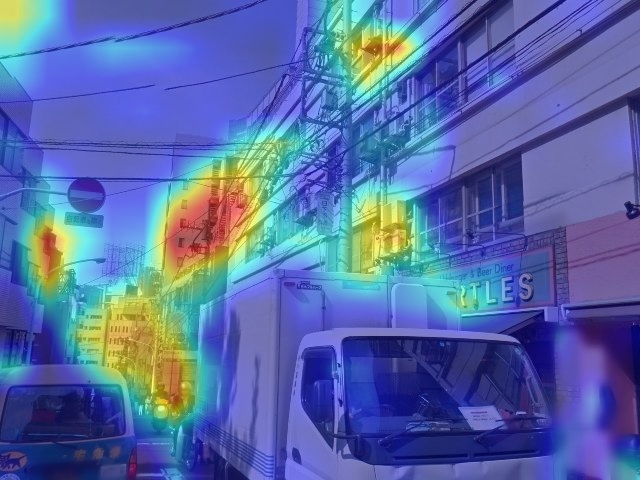}
    \vspace{-0.8\baselineskip}
\end{subfigure}
\begin{subfigure}[b]{0.11\textwidth}
\includegraphics[width=\textwidth,height=0.75\textwidth]{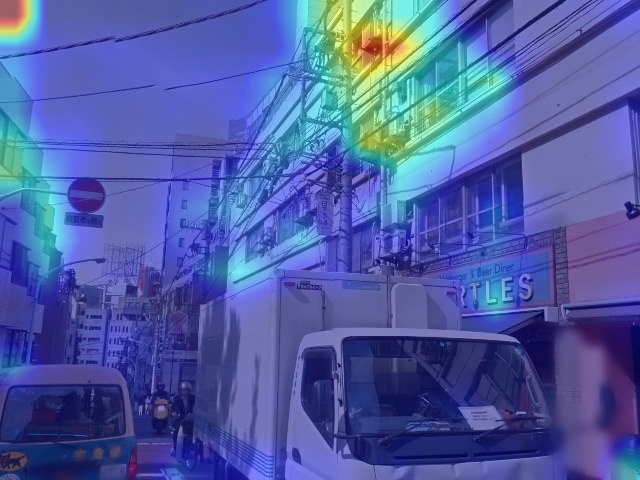}
    \vspace{-0.8\baselineskip}
\end{subfigure}
\begin{subfigure}[b]{0.11\textwidth}
\includegraphics[width=\textwidth,height=0.75\textwidth]{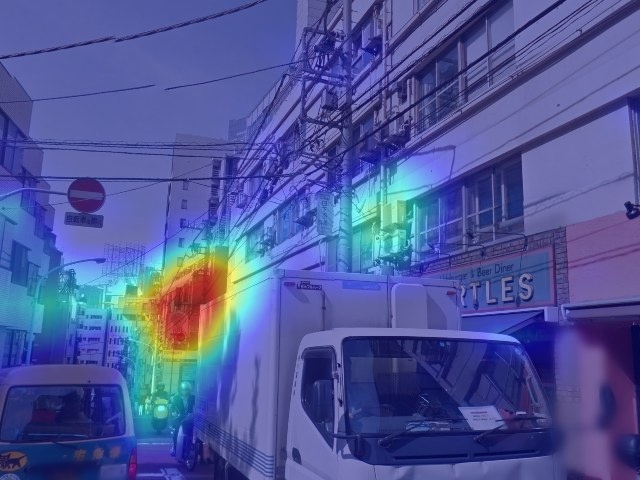}
    \vspace{-0.8\baselineskip}
\end{subfigure}
\begin{subfigure}[b]{0.11\textwidth}    \includegraphics[width=\textwidth,height=0.75\textwidth]{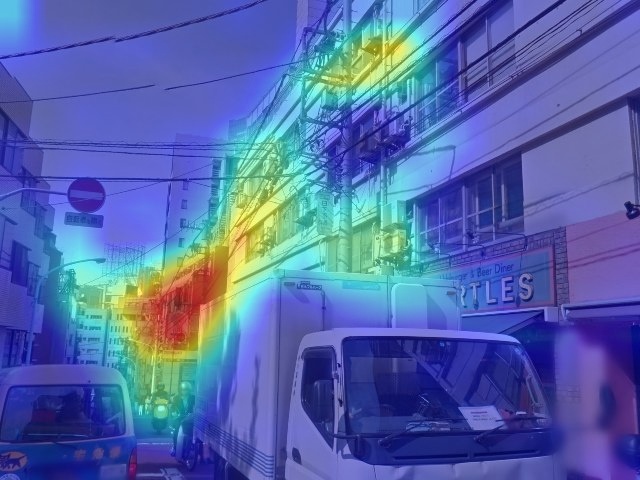}
    \vspace{-0.8\baselineskip}
\end{subfigure}
    \caption{The activation maps of ResNet-50 models trained with Triplet Loss (Baseline) and SSL methods. For each dataset, the first row is for the query image and the second row is for the positive sample image.}
    \label{act}
    \vspace{-10pt}
\end{figure*}

\subsubsection{Projected Embedding Dimensionality}

In examining the effect of projected embedding dimensions ($D_p$) on retrieval performance (Table \ref{ab}), we find a non-linear relationship between embedding size and metrics (R@1, R@5, R@10). SimCLR performs best at the lowest dimension (1024), with R@1 at 84.2\%, R@5 at 92.2\%, and R@10 at 94.3\%, while increasing $D_p$ to 2048 and 4096 slightly reduces performance.

In contrast, MoCov2, BYOL, SimSiam, and VICReg generally improve with larger $D_p$, with VICReg reaching its peak at 4096 (R@1: 77.7\%, R@5: 88.8\%, R@10: 90.7\%). BT, however, performs best at $D_p$ of 2048. This indicates that the optimal embedding dimension depends on the method and the interaction between model architecture and the complexity of the projected space.

\subsubsection{Feature Embeddings Dimensionality}

In Table \ref{ab}, we present the results for different feature embedding dimensions ($D_g$). The embedding dimension is key for memory consumption, inference and matching time in VG architectures, and it plays a critical role in improving model performance and representation ability.

Our data shows that SimCLR benefits from a larger $D_g$ (2048), with a notable increase in R@1, though R@5 and R@10 remain stable across dimensions. In contrast, MoCov2 performs best at lower $D_g$ values (512 and 1024), indicating less sensitivity to larger dimensions. BYOL, SimSiam, and VICReg favor the highest $D_g$ (2048), while BT's performance plateaus at $D_g = 1024$, showing diminishing returns beyond this point.

\subsection{Visualization}\label{vis_main}
In this study, we analyze activation maps from the final layers of ResNet-50 models trained with Triplet Loss (Baseline) and SSL methods, as shown in Fig.\ref{act}. Our findings reveal that, across all three datasets, models trained with Triplet Loss exhibit diffuse activation over irrelevant features, with inconsistent patterns under day-to-night shifts (MSLS) and viewpoint changes (Pitts30k). In the Tokyo24/7 dataset, these models mistakenly highlight dynamic objects like vehicles and roads. In contrast, our models, trained with SSL paradigm, focus more consistently on landmark regions and show uniform activation patterns. We also review the top-5 retrieval results of SSL methods on the MSLS validation set in Sec.\ref{visualize} of the appendix, demonstrating their effectiveness.

\subsection{Discussion}\label{dis}
The optimal training settings for various SSL methods (Table \ref{best}) underscore the importance of method-specific configurations to maximize performance. Our study highlights Hard Negative Mining (HNM), the number of projection layers, and the dimensions of projected and feature embeddings as key factors influencing geo-specific representations. We discuss HNM and projector settings by category and compare optimal SSL settings between general applications and VG, offering guidance for enhancing SSL in geo-related tasks.

\textbf{Contrastive Learning.} HNM improves SimCLR's performance but not MoCov2's, likely due to MoCov2's memory dictionary, which reduces the need for HNM. HNM may also reduce the diversity of MoCov2's memory bank. Additionally, multiple projection layers are less critical for both SimCLR and MoCov2 in VG than in general applications.

\textbf{Self-distillation Learning.} These methods show less reliance on HNM, likely due to their loss functions. Optimal performance is achieved with two projection layers and larger embedding dimensions, suggesting a preference for complex, high-dimensional embeddings. Notably, configurations differ between VG and general applications, with BYOL and SimSiam favoring different projector setups in VG. For example, BYOL prefers larger projected embeddings in VG compared to general application scenarios and SimSiam prefers a 2-layer projector instead of a 3-layer one.

\textbf{Information Maximization.} HNM benefits methods like Barlow Twins, which performs best with a 2-layer projection head and mid-range embedding sizes in VG, differing from its general application setup. VICReg achieves optimal results with three projection layers and large embedding dimensions, consistent with its general preferences.

\begin{table}[]
\centering
\caption{Optimal training settings for SSL methods in VG and general applications from original papers. For general applications, HNM is not applied, $D_{ph}$ is the hidden layer dimension of the projector, and $D_{po}$ is the output dimension of the projector.}\label{best}
\resizebox{1.0\linewidth}{!}{
\begin{tabular}{lccccccccc}
\toprule
    \multirow{2}{7em}{Categories} & \multirow{2}{4em}{Methods} & \multicolumn{3}{c}{VG} & & \multicolumn{3}{c}{General} \\
    \cline{3-5}\cline{7-9}
     & & HNM & $L$ & $D_p$ & & $L$ & $D_{ph}$ & $D_{po}$\\\midrule
    \multirow{2}{7em}{Contrastive Learning} & SimCLR  & \cmark & 1 & 1024 &  & 2 & 2048 & 128 \\
    & MoCov2  & & 0 & - & & 2 & 2048 & 128 \\\midrule
    \multirow{2}{7em}{Self-distillation Learning} & BYOL  &  & 2 & 4096 & & 2 & 4096 & 256 \\
    & SimSiam  & & 2 & 4096  & & 3 & 2048 & 2048 \\\midrule
    \multirow{2}{7em}{Information Maximization} & Barlow Twins  & \cmark & 2 & 2048 & & 3 & 8192 & 8192 \\
    & VICReg  & \cmark & 3 & 4096 & & 3 & 8192 & 8192 \\ 
    \bottomrule
\end{tabular}
}
\end{table}




\section{Conclusions}\label{sec:conclusions}
In this paper, we introduced the unified VG-SSL framework, which benchmarks various SSL methods alongside the GeoPair strategy for learning geo-specific representations in VG tasks. We evaluated six prominent SSL approaches for VG tasks in robotics and autonomous vehicles datasets, identifying optimal SSL settings, with a strong preference for contrastive learning and information maximization methods. SimCLR, MoCov2, and Barlow Twins emerged as top performers, matching or exceeding the performance of leading VG methods. Our findings emphasize the potential of SSL for VG tasks, and the flexibility of our framework enables its application to benchmark geo-specific representation learning problems. Additionally, the adaptable GeoPair strategy offers an opportunity extended to other retrieval-based tasks, such as face recognition. Future work will focus on integrating more SSL methods and exploring additional augmentations.

\noindent\textbf{Acknowledgement.} This work was supported by the Technology Innovation Institute, the NSF CAREER Award 2145277, the NSF CPS Grant CNS-2121391, and the NYU IT High Performance Computing resources, services, and staff expertise. Giuseppe Loianno serves as consultant for the Technology Innovation Institute. This arrangement has been reviewed and approved by the New York University in accordance with its policy on objectivity in research.

%
%
\bibliographystyle{splncs04}
\bibliography{egbib.bib}

\clearpage
\newpage
\appendix
\section{Appendix Overview}
This supplementary material provides the following additional information for a better understanding of our paper: Sec.~\ref{san} report our results on the Revisited San Francisco (R-SF) dataset. Sec.~\ref{limit} discusses the limitations of our work. Sec.~\ref{cat} summarizes the comparison of architecture for different SSL methods. Sec.~\ref{twostage} shows the extended results of our two-stage methods. Sec.~\ref{source} explains the data source of state-of-the-art VG results. Sec. \ref{preprocess} explains the preprocessing procedure in the inference stage. Sec~\ref{visualize} shows the analysis of visualized results of models trained with different SSL strategies.

\section{Results on R-SF}\label{san}
We add results on Revisited San Francisco (R-SF)~\cite{Chen_2011_san_francisco, RSF} in Table~\ref{san}, comparing with triplet loss baseline. The results reflect a similar trend as other datasets except that Barlow Twins does not show superior performance.

\section{Limitations}\label{limit}
This section reflects on the limitations of our research. A key limitation is the small batch sizes used (32 and 64), which are significantly smaller than those typically used in SSL methods, often exceeding 512. This may result in suboptimal performance. We chose smaller batch sizes to maintain comparability with other VG studies due to resource constraints. Larger batch sizes would have required more training steps to achieve similar results.

Another limitation is the scope of data augmentation. As noted in the Deep VG benchmark~\cite{Berton_CVPR_2022_benchmark}, the effectiveness of data augmentation can vary across datasets due to their unique characteristics. To establish a robust baseline, we focused on fundamental augmentations, such as random flipping and random resized cropping, which avoid dataset-specific biases and provide general benefits across all datasets. While we agree that exploring more advanced augmentations could enhance performance, applying them across multiple datasets can be complex and may result in inconsistent outcomes. We suggest investigating the impact of specific augmentations when fine-tuning models on individual datasets for more targeted improvements.

\begin{table}
\centering
\caption{Comparison of SSL methods and Triplet Loss on R-SF dataset}
\resizebox{\linewidth}{!}{
\begin{tabular}{lccc}
\toprule
    &
    \multicolumn{3}{c}{R-SF}\\
    \cline{2-4}
    & R@1 & R@5 &  R@10\\
    \midrule
    \multicolumn{1}{l}{\textit{Our One-Stage Methods with ResNet50-GeM}} \\
    Triplet Loss (Baseline) &44.6 & 58.0 & 63.2\\
    SimCLR & 46.8 & 64.0 & 69.1\\
    MoCov2 & 40.5 & 55.2 & 59.7\\
    BYOL & 25.1 & 38.1 & 46.0\\
    SimSiam & 25.3 & 39.3 & 44.6\\
    Barlow Twins & 22.4 & 36.0 & 41.6\\
    VICReg & 22.2 & 36.5 & 43.0\\\midrule
    \multicolumn{1}{l}{\textit{Our One-Stage Methods with DeiT-S}} \\
    Triplet Loss (Baseline) & 39.1 & 57.0 & 63.0\\
    SimCLR & 50.2 & 65.1 & 70.6\\
    MoCov2 & 35.3 & 53.8 & 58.2\\
    BYOL & 17.6 & 29.3 & 36.0\\
    SimSiam & 15.7 & 28.8 & 33.6\\
    Barlow Twins & 36.5 & 52.8 & 58.0 \\
    VICReg & 29.1 & 44.0 & 51.3\\
    \bottomrule
\end{tabular}
}
\end{table}

\section{Architecture difference of SSL methods}\label{cat}
Table \ref{related} presents a comparative analysis of various SSL methods, focusing specifically on their architectural differences. SSL techniques within identical categories tend to exhibit similarities in their loss functions and overall architectural designs. This observation encourages an analysis of our results based on categorical groupings. Nevertheless, variations in SSL approaches within the same category may lead to discrepancies in outcomes and the selection of optimal hyperparameters. For SimSiam~\cite{SimSiam}, we specifically note that we remove the BN layer for the output of the projection head since it does not converge with that BN layer.

\begin{table*}
    \centering
    \small
    \caption{Comparison of architecture for selected SSL methods. \textbf{ME}: Momentum target encoder. \textbf{SG}: Stop gradient for target encoder. \textbf{PR}: Predictor to infer target (teacher) embeddings based on online (student) embeddings. \textbf{BN}: Batch normalization in the projector or predictor. \textbf{LP}: Large dimensionality of projected embeddings.}
    \label{related}
    \begin{tabular}{cccccccc}
    \toprule
        Categories & Methods &  ME & SG & PR & BN & LP & Loss Function\\\midrule
         \multirow{2}{8em}{Contrastive Learning} & SimCLR \cite{simclr} & & & & & & InfoNCE Loss\\
         & MoCov2 \cite{mocov2} & \cmark & & & & &InfoNCE Loss\\\midrule
          \multirow{2}{8em}{Self-distillation Learning} & BYOL \cite{byol} & \cmark & \cmark& \cmark & \cmark & & Embedding Prediction Loss\\
         & SimSiam \cite{SimSiam} & & \cmark & \cmark & \cmark & & Embedding Prediction Loss\\\midrule
          \multirow{2}{8em}{Information Maximization} & Barlow Twins \cite{bt} & & & & \cmark&\cmark & Cross-correlation Loss\\
         & VICReg \cite{vicreg} & & & &\cmark&\cmark & VIC Regularization Loss\\\bottomrule
    \end{tabular}
\end{table*}

\section{Extended results of our two-stage methods}\label{twostage}
In Table \ref{twostageresult}, we show the extended results of the comparison between our two-stage methods with R2Former~\cite{r2former} for reranking and state-of-the-art two-stage methods (SP-SuperGlue~\cite{Sarlin_2020_CVPR}, Patch-NetVLAD~\cite{Hausler_2021_CVPR}, TransVPR~\cite{transvpr}, and R2Former~\cite{r2former}). For better readability, we concatenate the one-stage results (Table~\ref{sota}) at the bottom of the table.

When analyzing different SSL training strategies, it becomes evident that two-stage methods, particularly those rooted in contrastive learning and information maximization, demonstrate superior performance compared to self-distillation approaches. This trend aligns with observations made in one-stage results, underscoring the inherited high geo-specific representation quality from the first-stage results. Notably, among these strategies, SimCLR and Barlow Twins stand out, delivering higher overall performance metrics than their counterparts. 

In our comparative analysis of two-stage methods against leading approaches, we observed that SimCLR and Barlow Twins generally match or exceed the performance of existing state-of-the-art methods across most datasets, with the notable exception of the Tokyo24/7 dataset. Despite this, our enhancements to the original R2Former model yield only marginal gains, even though our initial-stage results surpass those of the original R2Former. To analyze the training bottleneck in two-stage methods, we conducted a detailed examination of the MSLS dataset's validation performance across various training stages, as outlined in Table~\ref{r2}.

We pay particular attention to the R@1 metric. In the global-retrieval-training phase, the performance ranking is as follows: SimCLR $>$ Barlow Twins$>$ R2Former $>$ MoCov2 $>$ VICReg $>$ BYOL $>$ SimSiam. However, this order shifts in the reranking-training phase to: SimCLR $>$ VICReg $>$ R2Former $>$ Barlow Twins $>$ MoCov2 $>$ SimSiam $>$ BYOL. This reshuffling illustrates a complex relationship between the outcomes of the global-retrieval-training and reranking-training stages, indicating that superior performance in the former does not automatically translate to enhanced results in the latter. 

Additionally, our findings reveal that post-finetuning, most variants reach a plateau in MSLS validation performance, with R@1 nearing 90\%, R@5 around 95\%, and R@10 close to 96\%. This saturation suggests a limit to the efficacy of the current fine-tuning approaches, highlighting the need for better strategies to push these metrics further.

\section{Data Source of State-of-the-art VG Results}\label{source}
In presenting the state-of-the-art results in VG methods as detailed in Table \ref{sota} and Table \ref{twostageresult}, we primarily draw upon data from several key research papers: R2Former \cite{r2former}, TransVPR \cite{transvpr}, Patch-NetVLAD \cite{Hausler_2021_CVPR}, and GCL \cite{Leyva-Vallina_2023_CVPR}. When considering the NetVLAD~\cite{netvlad} result, it is pertinent to acknowledge the variety of results yielded by different reproductions. For the purposes of this study, we refer to the results as documented by R2Former \cite{r2former}. However, it is notable that the results for the Nordland datasets and the dimension of the feature are not included in the paper, a gap attributed to the lack of clarity regarding the source of their reproduction. For the results of R2Former without reranking part, we download the model provided by the authors and evaluate it across VG datasets.

\section{Preprocessing for different datasets}\label{preprocess}
In the inference stage, due to the potential variability in the input images, preprocessing becomes an essential step. Our one-stage methods, as outlined in Table \ref{sota}, predominantly utilize resizing as a key preprocessing technique. This ensures that the dimensions of the input images align with those used during training. However, an exception is noted for the Tokyo24/7 dataset. Here, standard resizing procedures would alter the aspect ratios of the query images, potentially degrading performance. To address this, we adopt the \textit{single\_query} preprocessing approach, as described in \cite{Berton_CVPR_2022_benchmark}, specifically for the preprocessing of query images.

\section{Visualization results}\label{visualize}
\begin{figure*}
    \centering
    \includegraphics[width=\linewidth]{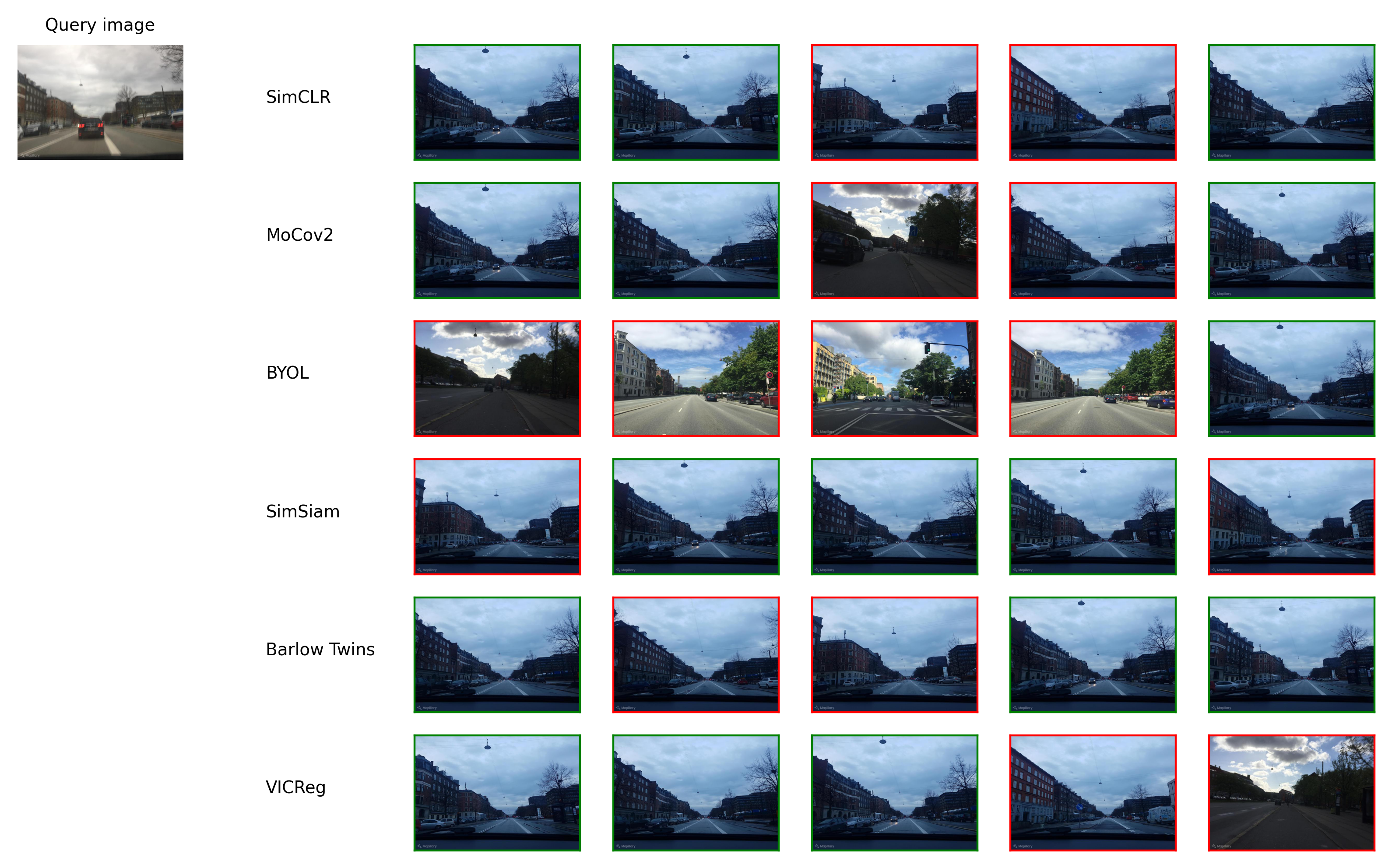}
    \caption{Visualization of top-5 retrieved candidates for \textbf{illumination change} across different SSL training strategies}
    \label{light}
\end{figure*}

\begin{figure*}
    \centering
    \includegraphics[width=\linewidth]{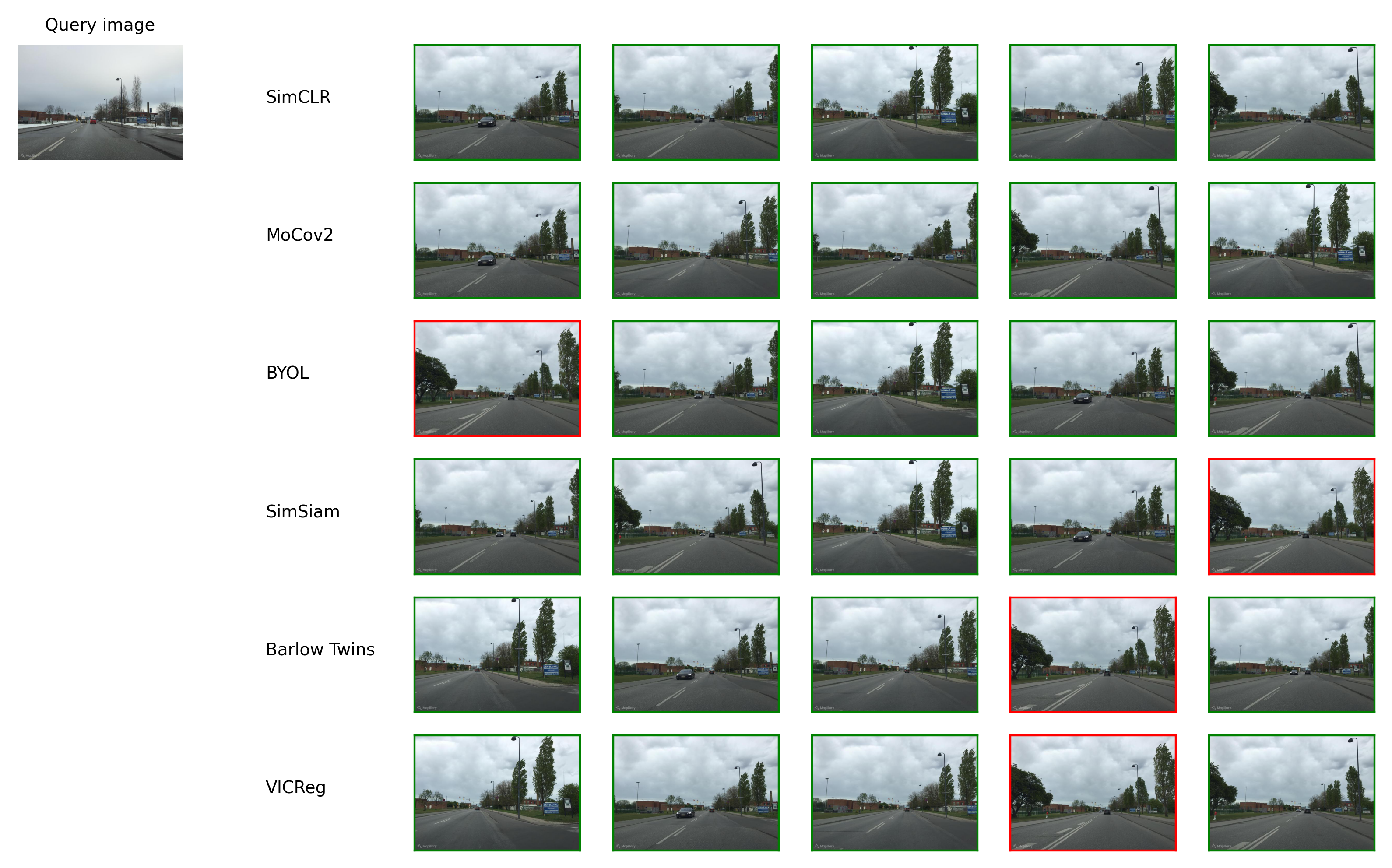}
    \caption{Visualization of top-5 retrieved candidates for  \textbf{season change} across different SSL training strategies}
    \label{season}
\end{figure*}

\begin{figure*}
    \centering
    \includegraphics[width=\linewidth]{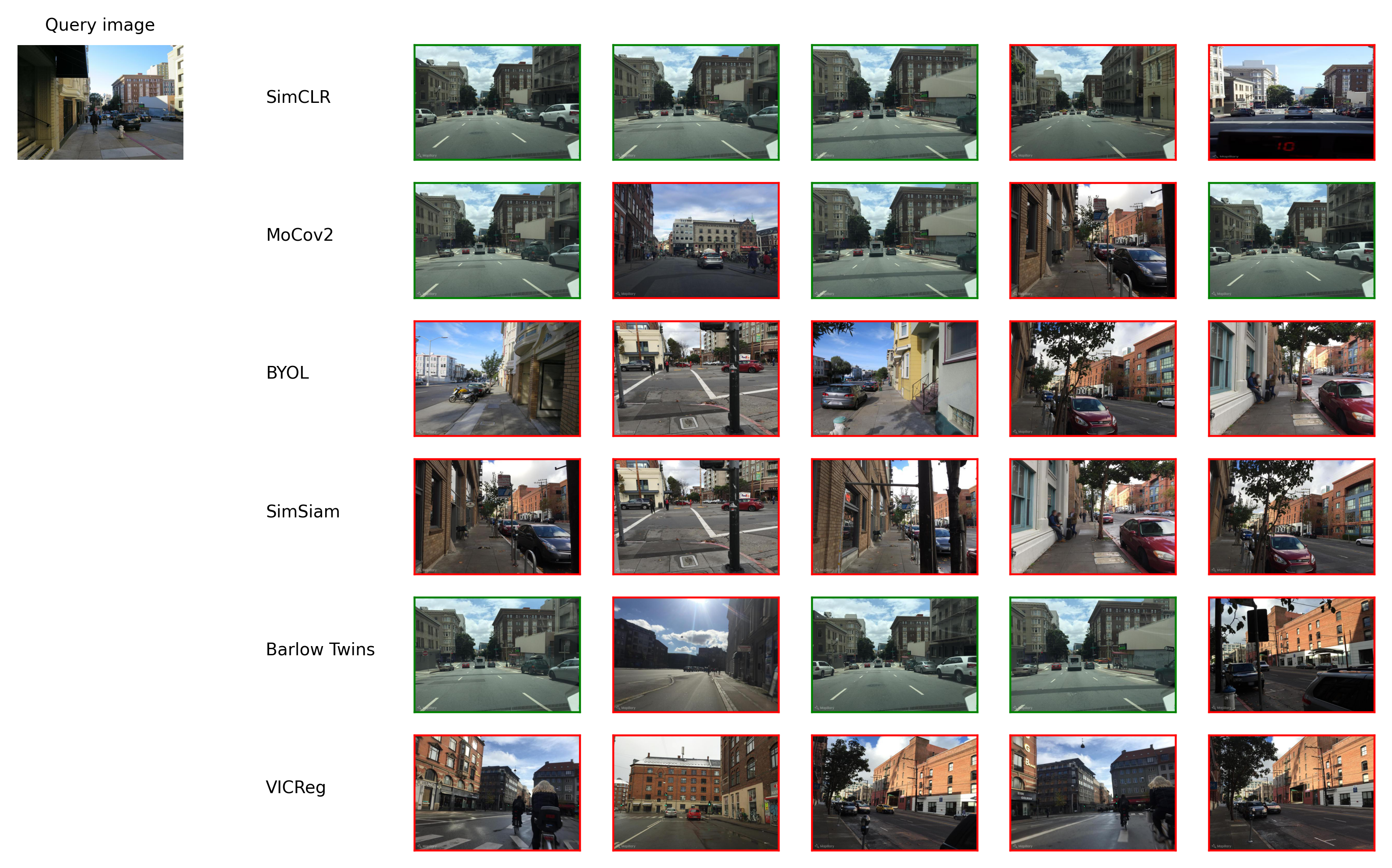}
    \caption{Visualization of top-5 retrieved candidates for  \textbf{viewpoint change} across different SSL training strategies}
    \label{viewpoint}
\end{figure*}

\begin{figure*}
    \centering
    \includegraphics[width=\linewidth]{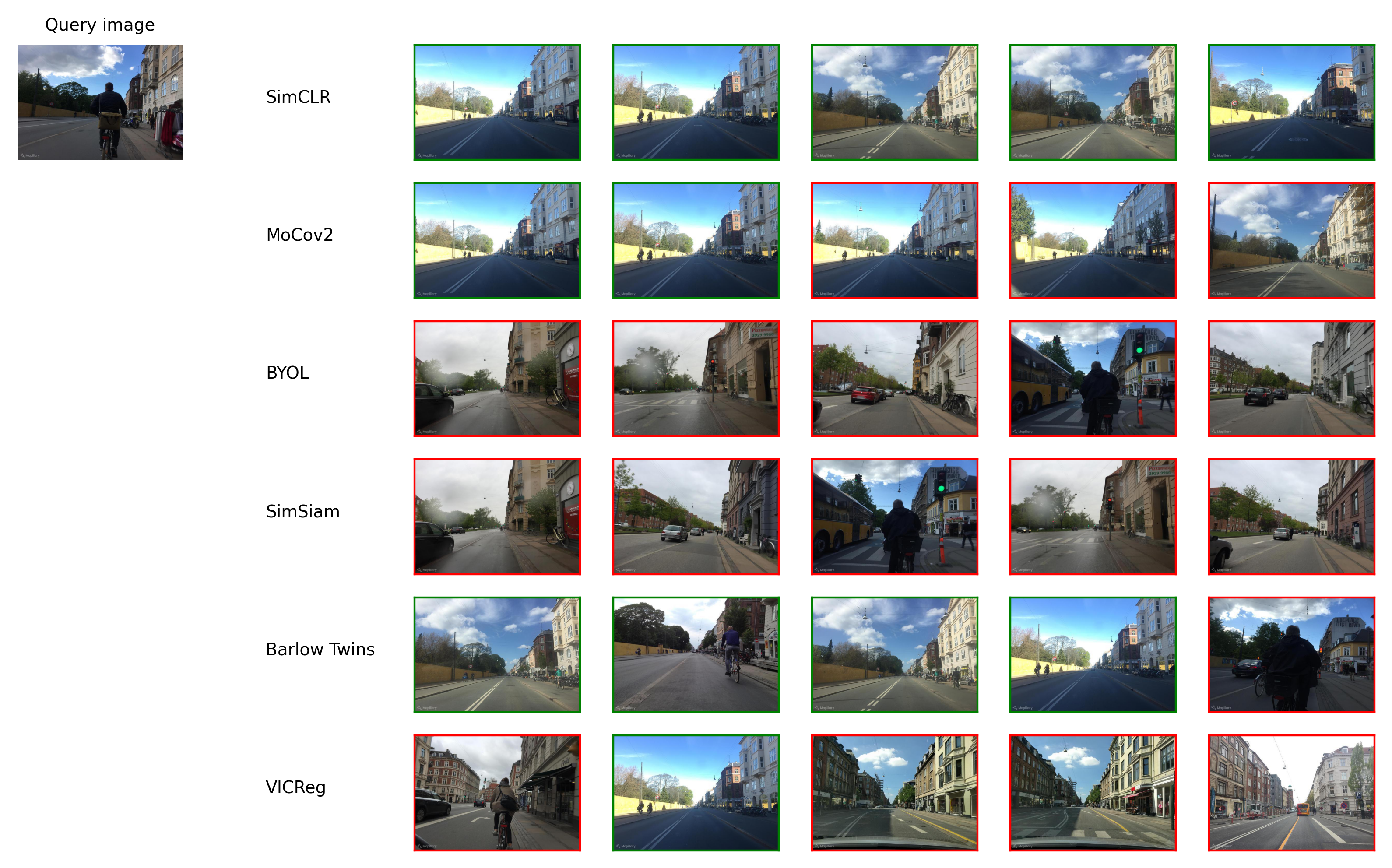}
    \caption{Visualization of top-5 retrieved candidates for  \textbf{occlusion} across different SSL training strategies}
    \label{occ}
\end{figure*}

In Fig.~\ref{light}~-~\ref{occ}, we present a comparative analysis of the top-5 retrieved images from the MSLS validation dataset, using ResNet50-GeM models trained via different self-supervised learning (SSL) strategies. This qualitative assessment specifically addresses challenges such as illumination change, seasonal change, viewpoint change, and occlusion. Our findings reveal that SimCLR, MoCov2, and Barlow Twins consistently outperform other methods in tackling these challenges, aligning with our quantitative results.

Notably, we observe that BYOL and SimSiam produce irrelevant outputs when confronted with changes in viewpoint and occlusion. This tendency may explain their lower recall performance, suggesting a deficiency in learning invariance against occlusion and viewpoint alteration. This insight is crucial as it highlights potential areas for refinement in these models, specifically in enhancing their robustness to such environmental and perspective shifts.

\begin{table*}
\centering
\caption{Comparison of state-of-the-art VG methods with our results on large-scale VG datasets. Our models were trained in the MSLS dataset. For the performance in the urban environment (Pitts30k and Tokyo24/7), we further finetuned our models in the Pitts30k dataset. The best results of one-stage and two-stage methods are with \textbf{bold} text separately, and the second and third best are \underline{underlined}. $^*$ shows the performance of the first stage without reranking. $^\dagger$ shows only the dimensionality of global embeddings but excludes local embeddings.}\label{twostageresult}
\scriptsize
\resizebox{\linewidth}{!}{
\begin{tabular}{lccccccccccccccccccccc}
\toprule
    &\multirow{2}{2em}{$D_g$}&
    \multicolumn{3}{c}{MSLS Val} & & \multicolumn{3}{c}{MSLS Challenge} & & \multicolumn{3}{c}{Pitts30k} & & \multicolumn{3}{c}{Tokyo24/7} & & \multicolumn{3}{c}{Nordland}\\
    \cline{3-5}\cline{7-9}\cline{11-13}\cline{15-17}\cline{19-21}
    & & R@1 & R@5 &  R@10&  & R@1 & R@5 &  R@10& & R@1 & R@5 &  R@10& & R@1 & R@5 &  R@10& & R@1 & R@5 &  R@10\\\midrule
    \multicolumn{4}{l}{\textit{Two-Stage Methods}} \\
    SP-SuperGlue~\cite{Sarlin_2020_CVPR} & - &  78.1& 81.9& 84.3& & 50.6 &  56.9&  58.3& & 87.2& 94.8&  \textbf{96.4}& &  \underline{88.2}& \underline{90.2}& \underline{90.2} & & 29.1 & 33.5 & 34.3\\
    Patch-NetVLAD~\cite{Hausler_2021_CVPR} & 4096$^\dagger$ & 79.5 & 86.2 & 87.7 & & 48.1 & 57.6 & 60.5 & & 88.7  & 94.5 & 95.9 &  & \underline{86.0} & \underline{88.6} & \underline{90.5} & & 44.9 & 50.2 & 52.2 \\
    TransVPR~\cite{transvpr} & 256$^\dagger$ & 86.8 & 91.2 & 92.4 & & 63.9 & 74.0 & 77.5 & & 89.0 &  94.9 & 96.2 & & 79.0 & 82.2 & 85.1& & \underline{58.8}& \textbf{75.0} & \textbf{78.7}	   \\
    R2Former~\cite{r2former} & 256$^\dagger$ & \underline{89.7} & \underline{95.0}& \textbf{96.2} & & \underline{73.0} & \underline{85.9}&  \textbf{88.8}& & \textbf{91.1}& \underline{95.2} & \underline{96.3}& & \textbf{88.6} & \textbf{91.4} & \textbf{91.7} & & \textbf{60.6}& \underline{66.8}& \underline{68.7}  \\\midrule
    \multicolumn{6}{l}{\textit{Our Two-Stage Methods with R2Former}} \\
    SimCLR & 256$^\dagger$ & \textbf{90.3} & \underline{95.0} & 95.4& &\underline{73.2} &\underline{85.6} &\underline{88.1} & &\underline{90.5} &\textbf{95.3} &\textbf{96.4} & &81.9 &87.3 &89.5 & &48.2  &54.3 &56.3    \\
    MoCov2 & 256$^\dagger$ & 87.4 & 93.1& 94.1& & 69.3 & 84.2 & 87.0 & &89.1 &\underline{94.9} &96.1 & &74.6 &82.2 &84.8 & & 38.7 & 47.2 & 50.0\\ 
    BYOL & 256$^\dagger$ & 87.2 &92.4 & 93.4& &69.1 &82.7 &84.7 & &88.3 &94.4 &95.7 & &78.1 &83.8 &86.3 & & 31.1 & 35.7 & 37.3\\
    SimSiam & 256$^\dagger$ & 86.2 & 94.2 & 95.1& &69.3 &82.5 &85.4 & &87.9 &94.1 &95.7 & &78.7 &83.5 &84.8 & & 45.0 & 50.6 & 52.5\\
    Barlow Twins & 256$^\dagger$ & 89.1 & \textbf{95.1} & \underline{95.9}& &\textbf{73.5} &\textbf{86.9} &\textbf{88.9} & &89.3 &94.6 &96.3 & &81.9 &87.0 &89.8 & & \underline{57.1} & \underline{65.6} & \underline{68.3} \\
    VICReg & 256$^\dagger$ & \underline{89.7}& 94.3 &\underline{96.1}& &72.6 &85.1 &\underline{88.1} & &\underline{89.4} &94.7 &96.2 & &77.5 &85.1 &87.0 & & 46.8 &55.1 & 57.9 \\
    \midrule
    \midrule
     \multicolumn{4}{l}{\textit{One-Stage Methods}} \\
    NetVLAD~\cite{netvlad} & - & 60.8 & 74.3 & 79.5 & &35.1 & 47.4 & 51.7& &81.9 &  91.2 & 93.7 & &\underline{64.8} &  \underline{78.4}& \underline{81.6} & & - & - & -   \\
    SFRS~\cite{ge2020self} & 4096&  69.2& 80.3 & 83.1& & 41.5& 52.0& 56.3& & \textbf{89.4}& \textbf{94.7}& \underline{95.9}& &  \textbf{85.4} &  \textbf{91.1} & \textbf{93.3} & & 18.8 & 32.8 & 39.8 \\
    TransVPR$^*$~\cite{transvpr} &  256 &70.8 & 85.1 &  89.6 & & 48.0 &  67.1 & 73.6 & & 73.8 &  88.1 & 91.9& & - & - &- & & 15.9 & 38.6 & 49.4\\
    R2Former$^*$~\cite{r2former} &  256 & 79.3 & 90.5 & 92.7 & & 54.9 & 75.1 & 79.6  & &72.9 & 88.5 & 92.6 & & 43.5 & 65.7 & 72.4 & &21.4 & 33.7 & 41.0  \\
    GCL-ResNet50-GeM~\cite{Leyva-Vallina_2023_CVPR} & 1024 & 74.6 &  84.7&  88.1 & &  52.9&  65.7&  71.9 & & 79.9 &   90.0& 92.8 & &  58.7& 71.1& 76.8 & & - & - & -    \\
    GCL-ResNeXt-GeM~\cite{Leyva-Vallina_2023_CVPR} &  1024 &80.9 & 90.7& 92.6 & & \underline{62.3}& \underline{76.2}& 81.1 & & 79.2 &  90.4& 93.2 & &  58.1& 74.3& 78.1 & & - & - & -    \\
    \midrule
    \multicolumn{4}{l}{\textit{Our One-Stage Methods with ResNet50-GeM}} \\
    SimCLR  & 1024 & \textbf{84.2} & \textbf{92.2} & \textbf{94.2} & & \textbf{63.1} & \textbf{78.9} & \textbf{83.6} & & \underline{82.8} & 91.9&  94.6 & & 54.6 & 74.9 & \underline{81.9}& & \textbf{39.9} & \textbf{56.4} & \textbf{63.9} \\
    MoCov2 & 1024 & \underline{81.5} & 90.5 & 92.8 & & 59.0 & 73.8 & 79.2& & 82.6 & \underline{92.4}&  95.1 & & 51.4 & 68.3 & 76.5 & & \underline{28.0}& \underline{42.7} & \underline{50.1} \\
    BYOL & 1024 & 72.7 & 85.5 & 87.7 & & 50.4 & 66.4 & 71.4 & & 80.2 &91.5 & 94.4 & &44.8 &63.8 &70.8 & &  10.6& 18.5 & 23.5 \\
    SimSiam & 1024 & 75.0 & 85.8 & 88.6 & & 52.1 & 67.0 & 72.2 & & 78.6 & 89.8 & 92.7 & &51.1 &67.6 &71.4 & &  12.5& 21.5 & 27.0  \\
    Barlow Twins & 1024 & 79.5 & 89.5 & 91.9 & & \underline{59.2} &74.2 & 79.1 & & 80.8 & 91.7 & 94.2 & &45.7 &61.9 &70.8 & &  18.5& 30.5 & 38.0  \\
    VICReg & 1024 & 77.4 & 89.3 & 91.2 & & 58.0 & 74.1 & 79.0 & & 80.2 & 91.3 & 94.1 & &50.2 & 65.4 &74.3 & &  14.9& 25.1 & 31.3  \\\midrule
    \multicolumn{6}{l}{\textit{Our One-Stage Methods with DeiT-S}} \\
    SimCLR & 256 & \underline{81.1} & \underline{91.1} & \underline{93.1} & &58.9 & \underline{77.1} & \underline{82.6}& &\underline{84.7} &\underline{93.9} &\textbf{96.0} & &\underline{59.4} &\underline{76.2} &80.0 & &24.9 & 38.9 & 46.1\\
    MoCov2 & 256 & 76.1 & 88.5 & 91.1& & 56.8 & 75.2 & 78.7& &80.8 &\underline{92.4} &95.0 & &50.8 &69.8 &77.1 & & 15.4 & 26.4 & 33.0 \\
    BYOL & 256 & 58.2 & 75.3 & 79.6 & & 37.7 & 54.0 & 60.4 & &76.6 &89.4&92.9 & &43.2 &62.2 &68.6 & &4.1 & 7.9 & 10.6\\
    SimSiam & 256 & 56.2 & 76.2& 80.1 & & 35.3 & 52.3 & 58.7 & &79.7 &91.0 &93.6 & &47.3 &63.8 &74.0 & & 6.2 & 11.5 & 15.4 \\
    Barlow Twins & 256 & 79.7 & \underline{91.4} & \underline{93.1}& & 59.1 & 76.1 & \underline{81.5}& &82.6 &92.1 &95.0 & &58.4 &75.2 &80.6 & & \underline{28.1} & \underline{43.3} & \underline{51.1} \\
    VICReg & 256 & 75.8 & 89.5 & 91.9 & & 56.9 & 74.0 & 78.2& &81.7 &92.3 &\underline{95.2} & &51.7 &66.7 &74.6 & &19.3  &32.1 &39.6  \\
    \bottomrule
\end{tabular}
}
\end{table*}

\begin{table*}
\centering
\caption{Comparison of validation performance in MSLS dataset for the different training stages of our two-stage methods.}\label{r2}
\resizebox{\linewidth}{!}{
\begin{tabular}{ccccccccccccccccccccccccccccc}
\toprule
      \multicolumn{3}{c}{R2Former~\cite{r2former}} & &\multicolumn{3}{c}{SimCLR} & & \multicolumn{3}{c}{MoCov2} & &\multicolumn{3}{c}{BYOL} & & \multicolumn{3}{c}{SimSiam}& &\multicolumn{3}{c}{Barlow Twins}& & \multicolumn{3}{c}{VICReg}\\
     \cline{1-3}\cline{5-7}\cline{9-11}\cline{13-15}\cline{17-19}\cline{21-23}\cline{25-27}
     R@1 & R@5 &  R@10& &R@1 & R@5 &  R@10&  & R@1 & R@5 &  R@10& & R@1 & R@5 &  R@10& & R@1 & R@5 & R@10 & &R@1 & R@5 &  R@10 & &R@1 & R@5 &  R@10\\\midrule
     \multicolumn{6}{l}{\textit{Global Retrieval Training}} \\
     79.3 & 90.8 & 92.6 & & 81.1 & 91.1 & 93.1 & & 76.1 & 88.5 & 91.1 & & 58.2 & 75.3 & 79.6 & & 56.2 & 76.2& 80.1 & &  79.7 & 91.4 & 93.1 & & 75.8 & 89.5 & 91.9\\\midrule
    \multicolumn{6}{l}{\textit{Reranking Training}} \\
     88.4 &  93.4 & 94.9 & & 89.2 & 94.3 & 95.4 & & 86.8 &93.1 & 93.9 & &81.2&86.2&87.4 & &83.0 & 87.3 & 88.8& & 88.2 & 92.7 & 93.6 & & 88.5 & 92.7 & 93.8\\\midrule
    \multicolumn{6}{l}{\textit{End-to-end Finetuning}} \\
    89.7 & 95.0 & 96.2 & &90.3 & 95.0 & 95.4& & 87.4 &93.1 &94.1 & &87.2 &92.4 &93.4 & &86.2 & 94.2 & 95.1 & &89.1& 95.1& 95.9& &89.7& 94.3 &96.1  \\ 
    \bottomrule
\end{tabular}
}
\end{table*}


\end{document}